\documentclass[journal]{IEEEtran}
\IEEEoverridecommandlockouts
\usepackage{caption}
\usepackage{subcaption}
\usepackage{comment}
\usepackage{amsmath,amssymb,amsfonts}
\usepackage{algorithm}
\usepackage[noend]{algpseudocode}
\usepackage{graphicx}
\usepackage{textcomp}
\usepackage{mathtools}
\usepackage{braket}
\usepackage{xcolor}
\usepackage{dirtytalk}
\usepackage{csquotes}
\usepackage{dsfont}
\usepackage{bbold}
\usepackage{multirow}
\usepackage{booktabs}
\usepackage{tabularx}
\usepackage{float}
\usepackage{pgfplots}
\pgfplotsset{compat=1.18}
\def\BibTeX{{\rm B\kern-.05em{\sc i\kern-.025em b}\kern-.08em
    T\kern-.1667em\lower.7ex\hbox{E}\kern-.125emX}}
\usepackage{nomencl}
\newcounter{mynum}
\newcommand{\nomentry}[2]{%
\stepcounter{mynum}%
\nomenclature[p\themynum]{#1}{#2}
}
\makenomenclature
\usepackage{cite}

\usepackage{scalerel}
\usepackage{tikz}
\usepackage{xcolor}
\usepackage{cancel}
\usepackage{placeins}

\usetikzlibrary{svg.path}

\definecolor{orcidlogocol}{HTML}{A6CE39}
\tikzset{
  orcidlogo/.pic={
    \fill[orcidlogocol] svg{M256,128c0,70.7-57.3,128-128,128C57.3,256,0,198.7,0,128C0,57.3,57.3,0,128,0C198.7,0,256,57.3,256,128z};
    \fill[white] svg{M86.3,186.2H70.9V79.1h15.4v48.4V186.2z}
                 svg{M108.9,79.1h41.6c39.6,0,57,28.3,57,53.6c0,27.5-21.5,53.6-56.8,53.6h-41.8V79.1z M124.3,172.4h24.5c34.9,0,42.9-26.5,42.9-39.7c0-21.5-13.7-39.7-43.7-39.7h-23.7V172.4z}
                 svg{M88.7,56.8c0,5.5-4.5,10.1-10.1,10.1c-5.6,0-10.1-4.6-10.1-10.1c0-5.6,4.5-10.1,10.1-10.1C84.2,46.7,88.7,51.3,88.7,56.8z};
  }
}

\newcommand\orcidicon[1]{\href{https://orcid.org/#1}{\mbox{\scalerel*{
\begin{tikzpicture}[yscale=-1,transform shape]
\pic{orcidlogo};
\end{tikzpicture}
}{|}}}}

\usepackage{hyperref} 

\begin{document}
\bstctlcite{IEEEexample:BSTcontrol}

\title{FedMap: Iterative Magnitude-Based Pruning for Communication-Efficient Federated Learning}

\author{
Alexander Herzog\IEEEauthorrefmark{1}\IEEEauthorrefmark{2} \orcidicon{0000-0003-1089-1815}, \and 
Robbie Southam\IEEEauthorrefmark{2} \orcidicon{0000-0002-0435-9980}, \and
Ioannis Mavromatis\IEEEauthorrefmark{2} \orcidicon{0000-0002-3309-132X}, \and
and \and
Aftab Khan\IEEEauthorrefmark{2} \orcidicon{0000-0002-3573-6240}\\
        \IEEEauthorblockA{\IEEEauthorrefmark{2}Toshiba Europe Ltd., Bristol Research \& Innovation Laboratory, UK}\\ 
        Email: \{Firstname.Lastname\}@toshiba-bril.com\\
        
}

        

\maketitle

\begin{abstract}
Federated Learning (FL) is a distributed machine learning approach that enables training on decentralized data while preserving privacy. However, FL systems often involve resource-constrained client devices with limited computational power, memory, storage, and bandwidth. This paper introduces FedMap, a novel method that aims to enhance the communication efficiency of FL deployments by collaboratively learning an increasingly sparse global model through iterative, unstructured pruning. Importantly, FedMap trains a global model from scratch, unlike other methods reported in the literature, making it ideal for privacy-critical use cases such as in the medical and finance domains, where suitable pre-training data is often limited. FedMap adapts iterative magnitude-based pruning to the FL setting, ensuring all clients prune and refine the same subset of the global model parameters, therefore gradually reducing the global model size and communication overhead. The iterative nature of FedMap, forming subsequent models as subsets of predecessors, avoids parameter reactivation issues seen in prior work, resulting in stable performance. In this paper we provide an extensive evaluation of FedMap across diverse settings, datasets, model architectures, and hyperparameters, assessing performance in both IID and non-IID environments. Comparative analysis against the baseline approach demonstrates FedMap's ability to achieve more stable client model performance. For IID scenarios, FedMap achieves over $90$\% pruning without significant performance degradation. In non-IID settings, it achieves at least $~80$\% pruning while maintaining accuracy. FedMap offers a promising solution to alleviate communication bottlenecks in FL systems while retaining model accuracy.
\end{abstract}


\section{Introduction}
Deep Neural Networks (DNNs) often require vast amounts of training data, particularly for intricate tasks such as image classification and language modelling.
In traditional, \emph{centralised}, Machine Learning (ML) approaches, data is typically brought together at a central location for training. 
This approach, however, encounters substantial challenges, particularly concerning data privacy regulations and GDPR issues (with associated fines for non-compliance) as well as the trustworthiness of the AI models~\cite{privacy_in_ml_challenges}. 

Federated Learning (FL)~\cite{Konecn2015FederatedOD} was proposed to address some of these challenges. 
It offers a framework for distributed training, primarily focusing on neural networks. 
Distinct from centralized models, FL ensures that raw data remains exclusively with the clients and is never transferred to the central server. 
In an archetypal FL system, data is harvested by client devices situated at the network's edge. 
The training regimen comprises local model updates using each client's distinct data and subsequently fusing all client models, usually via a central server or parameter server (PS). 
By this mechanism, not only is data privacy safeguarded -- a vital aspect in settings with privacy implications \cite{privacy_fl} -- but the model also offers an efficient solution to potential bandwidth constraints prevalent in large-scale use-cases, such as an Internet of Things (IoT) setting. 
Transmitting model parameters instead of raw data can significantly curtail data communication requirements \cite{10380759}.

However, it's crucial to note that client devices in an FL system frequently face resource limitations. 
Compared to the high-performance infrastructure in data centres, these client devices often exhibit reduced computational capabilities~\cite{fl_constrained_comp_heterogenous_devices_survey}, limited memory~\cite{fl_ressource_constrained_devices_survey, fl_iot_resource_constrained_perspective_survey}, smaller storage capacities~\cite{fl_io_medical_t_surveu}, and narrower communication bandwidth~\cite{comm_and_comp_eff_survey, overcoming_noisy_and_irrelevant_data, lim_future_aspects_comm_costs_fl, survey_fl_challenges_and_applications}. 
Our work particularly focuses on reducing communication bandwidth by collaboratively learning an increasingly sparse global model. 
To achieve this, we strictly adapt the iterative, magnitude-based pruning regime \cite{optimal_brain_damage} and apply it to the FL mechanism. 
Our method, named FedMap, ensures that all clients prune and refine using the same pruning mask, representing the same subset of the global model. 
As a result, once the client models are sufficiently refined, we proceed with further pruning to remove even more parameters.

Unlike prior approaches reported in the literature (e.g., PruneFL \cite{PruneFL} and FederatedPruning \cite{FederatedPruning}), a key advantage of FedMap is its ability to train a global model from scratch, making it well-suited for use cases with heightened privacy requirements. In the medical domain, increased sensitivity of healthcare data and stricter regulations make it difficult to leverage pre-trained models \cite{Rieke2020}, underscoring the value of our proposed method aiming to optimally train the global model without centralized data access. Similarly, in the finance sector, financial institutions are often -- and rightly so -- reluctant to share customer data due to privacy concerns, creating barriers to developing AI models that rely on pre-trained weights \cite{lee2023federated}. These challenges in such domains, where suitable pre-training data is often limited, position FedMap as a particularly advantageous solution compared to other methods reported in the literature.

Broadly, pruning techniques can be categorised into structured and unstructured~\cite{structured_vs_unstructured_pruning_survey}. 
While structured pruning \cite{strctured_pruning_survey} eliminates entire units like neurons or channels, unstructured pruning removes individual weights from the model -- 
FedMap is an unstructured pruning solution. As shown in the literature \cite{LAMP, pruning_for_compression_survey}, unstructured pruning has proven to achieve higher pruning ratios compared to structured pruning making it more suitable for communication-constrained FL deployments. Moreover, as all clients prune the same model and follow the same hyperparameters, masking information does not need to be exchanged. That means that the PS can reliably reconstruct the models and more importantly, that the models exchanged get ever-smaller after each round.

In this paper, we comprehensively explore FedMap's performance across a varied set of settings and pruning schedules, emphasising the versatility of our approach.
Moreover, we compare FedMap with \textit{FederatedPruning}~\cite{FederatedPruning}. A distinguishing feature of FedMap is its iterative approach, where subsequent models are consistently formed as subsets of their predecessors. 
This design precludes the parameter `reactivation' issue observed in \textit{FederatedPruning}; aiming for a more stable performance, particularly at higher pruning regimes. 
Through our experimental study, we show that FedMap contributes to a more stable client model performance, delivering superior results in both IID and non-IID settings.

Our main contributions in this paper are summarised below:
\begin{itemize}
    \item We propose a new method (FedMap) targeting the communication efficiency of FL deployments
    \item We extensively analyse our approach through a rigorous evaluation, with diverse settings (including different datasets, model architectures, hyperparameters), as well as evaluation under IID and non-IID scenarios for a realistic performance assessment.
    \item We also benchmark FedMap against the \textit{FederatedPruning} \cite{FederatedPruning} approach and successfully demonstrate the effectiveness of our approach.
\end{itemize}

This paper is organized as follows: Section \ref{sec:rel_work} covers preliminaries and related work, including Federated Learning, Magnitude-Based Pruning, and relevant literature. 
Section \ref{sec:method} introduces our method, \textit{FedMap}, and its extension for heterogeneous data distributions. Section \ref{sec:experiments} details our experiments, including various pruning schedules, local training influences, non-IID data effects, and benchmarking. Section \ref{sec:results_disc} discusses the results, focusing on model performance, stability, and the integration of FedDR for the deployment of FedMap to non-IID settings. Section \ref{sec:concl_lim_fut} concludes with a summary of findings, limitations, and future research directions. Additional results are provided in the appendices.

\section{Preliminaries and Related Work}
\label{sec:rel_work}
Before further explaining FedMap's functionality, we present integral details of FL and magnitude-based pruning, setting a solid foundation for further analysis. We also briefly compare FedMap with other similar solutions found in the literature, paving the way for our experimental comparison in the following sections. 

\subsection{Federated Learning}
Federated learning (FL) is a collaborative, privacy-preserving method for training machine learning models in a distributed manner. 
During each iteration of training/FL round $t \in \{1, ..., T\}$ , a client $n \in \{1, ..., N\}$, with $T,N \in \mathbb{N}^*$ utilise a subset of their local data $\mathcal{D}_{n,t}$, i.e., ${B_{n, t} \subset \mathcal{D}_{n, t}}$ (where ${B_{n, t}}$ denotes a batch of training examples), to calculate gradients with respect to the global model's parameters, $\theta_t \in \mathbb{R}^d$:
\begin{equation}
    \nabla_{n, t, B_{n}} = \frac{\partial \mathcal{L}(\theta_{t}, B_{n, t})}{\partial \theta_{t}},
\end{equation}
where $\mathcal{L}$ denotes the loss function. 
The choice of consensus mechanism within FL influences whether gradients revert to the PS for aggregation, e.g., Federated Stochastic Gradient Descent (FedSGD) method of aggregation \cite{Konecn2015FederatedOD} or Federated Averaging \cite{FedAvg}. 


Federated Averaging (FedAvg) \cite{FedAvg}, one of the most prevalent aggregation approaches, employs locally computed gradients across multiple batches contained in the client $n$'s training set at FL iteration $t$; ${\mathcal{D}_{n, t} = \{B_{n, t}^{1}, ..., B_{n, t}^{\tau}\}}$ with $\tau = |D_{n, t}|$ and  ${\mathcal{D}_{n, t} \subseteq \mathcal{D}_n}$. 

Subsequently, the deviation between the updated local model and the global model parameters, defined as $\Delta \theta_{n, t} = \theta_{t} - \theta_{n,t}$ is communicated to the PS. By employing multiple batches and facilitating local updates, FedAvg optimizes communication rounds between the PS and clients, effectively reducing the communication frequency.
The strategy the PS adopts for aggregation determines the formation of the subsequent global model. In the process of FedAvg, the changes in model parameters from multiple client models are aggregated in an element-wise manner to compute a global model update:
\begin{equation}
\theta_{t+1} = \theta_t + \frac{1}{N}\cdot\sum^{N}_{n=1}\Delta{\theta}_{n, t},
\label{eq:fed_avg}
\end{equation}

It is crucial to recognise that variations in data collection across clients and temporal shifts as during FL training rounds may impact model performance. 
In cases of independently and identically distributed client data (IID), the model parameters exhibit significant adjustments in the initial FL rounds, which progressively decrease. However, in practical scenarios, data leans toward non-IID \cite{non_iid_decentr_ml}, implying various data-distribution shifts. 
In this work, we consider both IID and non-IID scenarios, evaluating our proposed FedMap methodology considering feature-, label- and sample-heterogeneity. See \cite{sota_survey_non_iid_methods_fl} for a more detailed discussion of statistical heterogeneity in FL settings.

\subsection{Magnitude-Based Pruning}
\label{subsec:imap}
Achieving a high level of compression for neural networks can be accomplished by eliminating parameters that have a minimal effect on model performance. 
In unstructured pruning, the deletion of parameters is achieved by setting the values of the parameters and the gradients during back-propagation to $0$ \cite{structured_vs_unstructured_pruning_survey}. 
The importance of parameters can be assessed according to the impact of removing a parameter on validation loss $\mathcal{L}(\theta_{t}, \mathcal{D}_{Val})$. 
Parameters whose elimination incurs small perturbations to the validation loss are prioritised for removal. 
Although there is a rich body of work on pruning at initialisation \cite{SNIP, IterSNIP, SynFlow, GrasP, PHEW}, iterative pruning has a proven track record of learning highly sparse neural networks while maintaining a performance comparable to its densely parameterised counterparts. The iterative pruning routine was first described as early as 1989 in the work of LeCun et al. \cite{optimal_brain_damage} proposing Optimal Brain Damage (OBD). 

Apart from the amendments (italic font) as result of later findings \cite{Castellano1997AnIP, Mocanu2018, LAMP} iterative pruning in its current form has largely been described in the OBD procedure:
\begin{enumerate}
    \item Choose a suitable network architecture
    \item Train the network until a reasonable solution is obtained.
    \item \textit{Determine the layerwise sparsity levels $p_{i}$ according to a pre-selected pruning paradigm.}
    \item Sort the parameters \textit{according to their magnitude} and in every layer $i$ delete the smallest $1 - p_{i}$ (according to 3) parameters by magnitude.
    \item Iterate to step 2
\end{enumerate}
Iterative magnitude-based pruning with layerwise pruning ratios is described in Algorithm \ref{alg:imp}.

\begin{algorithm}[t]
\caption{Iterative Layerwise Magnitude-Based Pruning}
\label{alg:imp}
\begin{algorithmic}[1]
\Require{The untrained neural network, randomly initialised with parameters $\theta$; The global pruning rate $p_G$; The number of iterations $T$; The number of fine-tuning iterations $\mathbf{s}$}
\For{$t=1$ to $T$}
    \State Train the network for $L$ epochs.
    \If{t \% $\mathbf{s}$  == 0}
        \State Compute the layerwise pruning ratios $p_i$ according \Statex \hspace*{2.6em} to layerwise, magnitude-based pruning scheme,
        \Statex \hspace*{2.6em} s.t. $\sum_{i=1}^{L} p_i |W^i| =  p_G \cdot d,\quad \forall W^i \in \theta$.
        \State Sort by the magnitude and set all but the top 
        \Statex \hspace*{2.6em} $(1-p_i)$ \% of $W_i$ to 0.
    \EndIf
\EndFor
\Return{The pruned network $\Tilde{\theta}$.}
\end{algorithmic}
\end{algorithm}

It is important to note that the selection of parameters $s$ and $L$ significantly affects the model's convergence behaviour, especially during the first iteration of the algorithm. 
Therefore, careful selection of these parameters is important to ensure that the model is trained towards a good initial solution (good validation set performance) before proceeding with pruning. 

Since the inception of iterative pruning in \cite{optimal_brain_damage}, many methods have been proposed to determine the importance of parameters. 
More recent work hints at the effectiveness of simple magnitude-based pruning (\textit{MaP}) \cite{state_of_sparsity_in_nns}, where the magnitude of the parameters is used as a proxy to determine importance. 
The removal of parameters smallest in magnitude is assumed to have the smallest impact on $\mathcal{L}(\theta_{t}, \mathcal{D}_{Test})$. 

Using parameter magnitude as a proxy for importance can be explained by viewing \textit{MaP} as relaxed layerwise $l_2$ distortion minimization  \cite{LAMP}. 
Without considering the activation function $\sigma$, NNs can be viewed as linear operators in layers $W^{i} \in \mathbb{R}^{m \times n \times [p] \times [q]}$, where $W^{i}$ is the weight tensor of layer $i$, acting as an operator on input vector $x$. 
Given $W^{i} \in \mathbb{R}^{m \times n}$ and $x \in \mathbb{R}^n$, the induced matrix $l$ norm measures how much $W^{i}$ affects the length of $x$, denoted as:
\begin{equation}
    ||W^{i}||_{l} = \sup_{x \neq 0}{\frac{||W^{i}x||_{l}}{||x||_l}}.
\end{equation}

Viewing neural network layers as nested linear operators have been the basis of recent pruning methodologies such as 
in the layer-adaptive magnitude-based pruning (LAMP) scheme \cite{LAMP}. 
The LAMP score is a re-scaled version of the magnitude that incorporates the model-level $l_2$-distortion incurred by pruning. 
Given a global pruning target ratio $p_G$, the LAMP score determines layerwise pruning levels, $p_i$, while the magnitude of the parameter determines the inclusion or exclusion in the set of prunables. 
Compared to other common selection schemes for MP such as the Erdős–Rényi-Kernel methodology \cite{Mocanu2018, RiGL}, layerwise- ($p_i = p_G$) or global (select $p_G$ \% of parameters uniformly across all $L$ layers), the LAMP score marks the state of the art for iterative magnitude-based pruning.

\subsection{Related Work}
\textbf{Communication Efficiency}: Efficient communication is a significant bottleneck in FL. 
Standard solutions to prevent excessive communication overheads include reducing the update frequency between the server and the clients or size of the model updates \cite{10380759}. 
Many methods exist that precisely target communication efficiency, ranging from quantisation to standard compression techniques (e.g. finite source coding) or a combination of both. 
Traditional sparsification methods such as \textit{Top-K} sparsification-based methods achieve excellent performance even under strict compression regimes. Additionally, their convergence has theoretically been proven \cite{convergence_sparse_grad_methods}, especially in combination with error-accumulation \cite{sparse_sgd_memory}. 
In \textit{Top-K}, parameters are selected via the magnitude of the difference between the global- and the local client model. The selected value-index pairs are then sent to the server. 
Typically, the parameter's floating value is compressed by quantisation, and the integer encoding of the index can be compressed via Golomb coding \cite{sparse_bin_comp}. 
There is a rich body of work on reducing or eliminating the need to send positional information. 
The \textit{SmartIdx} method by Wu et al. \cite{SmartIdx}, is one example. 
The authors propose a structured pruning scheme, where only whole structures, such as convolutional filters and their module index, are sent, reducing the sending indices from 1:1 to n:1, where $n$ is the number of parameters in the substructure. 
Another prominent example is the \textit{TCS} method by Ozfatura et al. \cite{TCS}. 
The authors extend the canonical \textit{Top-K} method with error accumulation and reduce the number of non-zero entries in the binary index mask by reusing the previously obtained global mask to select the client parameters plus a small percentage of additional parameters.


\textbf{Sparsification of Neural Networks}:
Sparsification or pruning (i.e. the removal of connections by setting the respective weights to zero) is a broad term to describe the process of finding subnetworks with similar- or (in some cases) better performance than their dense counterparts. 
The most common are dense-to-sparse methodologies, where a dense network is gradually pruned throughout the training process. 
Recent discoveries on neural network pruning reveal that with carefully chosen layerwise sparsity, a simple magnitude-based can achieve a state-of-the-art tradeoff between sparsity and performance. Many methods resort to handcrafted heuristics \cite{Mocanu2018}, \cite{RiGL} such as keeping parts of the model dense, including the first or last layers \cite{dynamic_model_pruning_with_feedback}, \cite{dynamic_sparse_reparam}, \cite{to_prune_not_to_prune}. Kusupati et al. \cite{soft_thresh_weight_reparam} provides an overview of the different pruning methods. 

\textit{PruneFL} \cite{PruneFL} method of pruning under a FL framework outlines a two-stage pruning process. 
In the initial stage, a powerful and trusted client pretrains the model while also adaptively pruning the model until the model size stabilises.
This approach ensures a reduced computation and communication time for each federated learning round, especially in heterogeneous setups. However, the initial pruning might not be the most efficient since it draws from the data of just one client. 
To circumvent this, further adaptive pruning is introduced during FL, employing the standard \textit{FedAvg} procedure and engaging data from all participating clients. 
Their adaptive pruning involves removing and adding parameters, which the authors call `reconfiguration'. 
Notably, this reconfiguration transpires during multiple FL rounds at the selected client or the server post-client updates. 
A significant element of their method is the communication benefit, where bitmaps efficiently communicate client pruning details in the uplink.
In contrast, our proposed \textit{FedMap} method aligns more with Iterative Magnitude-Based Pruning. 
Instead of following a pre-training (and simultaneous pruning) procedure as in \textit{PruneFL}, we begin with an untrained model, aiming to collaboratively learning a sparse global model. We emphasise client-side pruning, avoiding mask transmissions and ensuring predictable communication overheads. We introduce flexibility in pruning schedules without broadcasting bitmaps during the uplink. 

Similarlily, \textit{Federated Pruning} \cite{FederatedPruning}  begins each round with the server crafting a set of variable masks and dispatching the pruned model to clients. 
This strategy shares a common thread -- pruning at the round's onset. 
However, while the pruning masks are crafted server-side, in our proposed setup, clients prune models on their end, eliminating the need for transmitting a sparse model. 
To ensure reproducibility, our clients and the Parameter Server (PS) initialise their models using the same random seed. 
While \textit{Federated Pruning} resorts to structured pruning, our approach, reminiscent of \textit{PruneFL}, opts for unstructured pruning.

In summary, our proposed approach, FedMap, offers several innovative advantages over existing federated learning and pruning methods like PruneFL and Federated Pruning. 
It features an analogous implementation to iterative magnitude-based pruning, collaboratively learning a sparse global model from dense client models without pretraining. 
Critically, it involves clients in the pruning process before training, generating model masks locally to negate transmission of masking information or sparse models, ensuring predictable communication costs. 
Unlike prior work that deflates models before aggregation, FedMap performs aggregation directly on dense client updates when following the same pruning schedule. 
It offers flexible, client-driven pruning schedules without incurring overhead from bitmap transmissions. 

\section{Iterative Magnitude-based pruning for compression in FL}
\label{sec:method}
The Lottery Ticket Hypothesis (LTH) \cite{lottery_ticket_hypothesis} states that given a large neural network and a specific training task; there exists a sub-network within it (``the winning ticket"), indicated by the nonzero elements of a binary mask $M \in \mathbb{R}^d$, that when trained from its initialisation, can achieve comparable or even better performance than the full network, using a fraction of the parameters, $||M||_0 << d$. 
This observation has led researchers to investigate techniques for finding such sub-networks, aiming to reduce the computational resources required for training and inference. 
In the context of FL, Lottery Ticket Networks (LTN) are commonly determined by pruning the global model using client data. Most prominent in this context is the \textit{LotteryFL} method \cite{LotteryFL}, where each client learns a personalised LTN (i.e. a sub-network of the base model). 
Therefore, client-server communication can be drastically reduced due to the compact size of the lottery networks. 
Since the LTNs are personalised, both the values and their positional information are communicated back to the server. 

The observation of a similar test accuracy between the pruned and unpruned network in a FL setup suggests the existence of a sparse, lottery ticket-like sub-network, which can be uncovered during training.
The insights gained from this observation offer two avenues to reduce communication overheads:
\begin{enumerate}
    \item When clients prune after training on their local datasets $\mathcal{D}_{n, t}$, clients need to transmit  $||M_{n, t} \odot \theta_{n, t}||_0$ to the PS in addition to $|M_{n, t}| = d$ bits to signal the pruned locations.
    \item Clients prune the global model before they begin training on their individual datasets. Given that the parameter server already has access to the global model, the clients are only required to send the parameters described by  $||M_{t} \odot \theta_{n, t}||_0$. Here, $M_t$ is the pruning mask determined by the selected pruning technique, applied to the global model parameters before a client's training -- which is the same on all client devices.
\end{enumerate}

Our study is primarily focused on the latter approach, which offers some distinct benefits.
In the context of approach 1, each client produces a unique pruning mask, this results in a collection of distinct pruning masks $\mathbf{M} = \{M_1, M_2, ..., M_n\}$ for the $N$ clients post-training. 
For any two clients, $i$ and $j$ from the set of all clients, the support of $M_i$ may or may not be a subset of the support of $M_j$. 
Consequently, the union of supports of all masks is always greater than or equal to $(1-p_G) \cdot d$. Such a scenario can be interpreted as a form of personalisation, similar to what has been demonstrated in \cite{FedMask}. 
On the PS, client updates are aggregated according to their overlapping pruning masks
to produce a new global model. This demands sending more parameters and additionally requires transmitting the merged client mask, resulting in an enlarged downlink payload.

Contrarily, with approach 2, all clients share a common support, leading to a uniformity in the masks; $supp(M_i) = supp(M_j)$, hence $M_{i, t} = M_{j, t}$, ensuring the union of mask supports is precisely $|\bigcup^{n}_{i = 1} supp(M_i)| = (1-p_G) \cdot d$, as long as all clients employ the same pruning rate $p_G$.
We choose our pruning schedules s.t. clients train unpruned models during the initial phase of FL training, similar to what is demanded as initial condition in Algorithm \ref{alg:imp}. 
Post this phase, pruning is executed before clients retrain their models, analogous to lines 2, 3 and 4 in Algorithm \ref{alg:imp}. 

\subsection{\textit{FedMap} Design principle}
The main design principle behind \textit{FedMap} revolves around collaboratively learning and sparsifying the global model. 
In accordance with the iterative pruning methodology (see Section \ref{subsec:imap}), clients follow a pruning schedule, which determines \underline{how} (we utilise the pruning methodology outlined in \cite{LAMP}) and \underline{when} (the size of the pruning interval $\mathbf{s}$, measured in FL rounds $t$) 
to prune global model parameters. 
After clients locally update the global models for $L$ local epochs on their respective datasets $D_{n, t}$; clients compress the model by only keeping all nonzero values of the sparse client-model, denoted as $\Delta\check{\theta}_{n, t} \in \mathbb{R}^{\lfloor (1 - p_G) \cdot d \rceil}$, which is then communicated to the PS, where aggregation is performed as follows:
\begin{equation}
    \Delta\Tilde{\theta}_{t+1} = \frac{\sum_{n=1}^{N} \Delta\Tilde{\theta}_{n, t} \cdot \mathbb{1}(\Delta\Tilde{\theta}_{n, t} \neq 0)}{\sum_{n=1}^{N} \mathbb{1}(\Delta\Tilde{\theta}_{n, t} \neq 0)}. 
    \label{eq:global_averaging}
\end{equation}

Producing the new global model is achieved by expanding the aggregation method proposed in \cite{FedAvg} (see Equation \ref{eq:fed_avg}), to account for the client-subsets. 
Specifically, for each client $n \in N$, the algorithm accumulates the changes in parameters $\Delta\Tilde{\theta}_{n, t}$ only when they are non-zero, denoted via the indicator function $\mathbb{1}(\Delta\Tilde{\theta}_{n, t} \neq 0)$. 
The summation of all these non-zero changes is then normalised by the total number of clients that had non-zero changes at these specific locations in their weights. 
In essence, the algorithm computes a weighted average of parameter changes across all clients, depending on the support of individual client masks. 

At the beginning of each round $t$, each client $n$ receives the global update $\Delta\check{\theta}_t$ from the PS (see Algorithm \ref{alg:fedmap}). 
Since all clients follow the same pruning schedule, the sparse version of the update can be recovered using the pruning mask from the previous round, $M_{t-1}$, to obtain $\Delta\Tilde{\theta}_t$. 
This operation is abbreviated with the function $RFM(\cdot)$ (\textbf{R}ecover \textbf{F}rom \textbf{M}ask), which can be formalised as follows: 
\begin{equation}
    \Tilde{\theta} = \left\{\begin{matrix}
\check{\theta}_{f(i)} & \text{ if }M_i = 1\\ 
0 & \text{otherwise,}
\end{matrix}\right.
\label{eq:rfm}
\end{equation}
where $f(i)$ gives the index in $\check{\theta}$ corresponding to the $i$-th non-zero entry in $M$. 

After recovering the global model from the recovered model-delta $\Delta\Tilde{\theta}_t$ (see line 6 in Algorithm \ref{alg:fedmap}), the client computes the size of the new set of nonzero parameters $K_t \in \mathbb{R}^{0:d}$ as a function of the current FL round $t$ and model-size $d$ according to the pruning schedule, denoted as $Schedule(\cdot)$. 
For a more thorough description of the pruning schedules, see Section \ref{subsec:pruning_schedules}. The function $Prune(\cdot)$ prunes the model to $K_t$ non-zero parameters; returning the pruned model $\theta_t$ and the corresponding pruning mask $M_t$. 
We use the aforementioned \textit{LAMP} method \cite{LAMP} in the context of this work. 
After training the sparse global model on $\mathcal{D}_{n,t}$ to obtain $\Tilde{\theta}_{n, t}$, the residuals are compressed via the $RWZ(\cdot)$ (\textbf{R}emove \textbf{W}here \textbf{Z}ero) function. 
This function eliminates all sparse locations in $\Delta\Tilde{\theta}_{n, t}$ and returns $\Delta\check{\theta}_{n, t} \in \mathbb{R}^{K_t}$.

The communication benefit with \textit{FedMap} stems from only ever communicating $K_t$ parameters during the uplink (see lines 12 and 17 in Algorithm \ref{alg:fedmap}), without the necessity of sending additional masking information.
In \textit{FedMap}, the PS acts solely as a vendor of aggregated client updates (see lines 16 and 17 in Algorithm \ref{alg:fedmap}), which has the same communication advantage for the downlink.

\begin{algorithm}[t] 
    \caption{The FedMap Algorithm}
    \label{alg:fedmap}
    \begin{algorithmic}[1]
        \For{$t=1,...,T$}
            \State \underline{\textbf{Client Side:}}
            \For{$n=1,...,N$}
                \State Receive $\Delta{\check{\theta}_{t}}$ from the PS 
                \State $\Delta \Tilde{\theta}_{t} = RFM\left(\Delta\check{\theta}_{t}, M_{t-1}\right)$
                \State {Update Global Model: $\Tilde{\theta}_t = \Tilde{\theta}_{t-1} - \Delta \Tilde{\theta}_{t}$}
                \State $K_{t} = Schedule(d, t)$
                \State $\Tilde{\theta}_{t}, M_{t} \leftarrow Prune(\Tilde{\theta}_t, K_{t})$
                \State Train $\Tilde{\theta}_t$ on $\mathcal{D}_{n, t}$ for $L$ epochs to obtain $\Tilde{\theta}_{n, t}$.
                \State $\Delta \Tilde{\theta}_{n, t} = \Tilde{\theta}_{t} - \Tilde{\theta}_{n, t}$
                \State {Compress: $\Delta \check{\theta}_{n, t} = RWZ\left(\Delta \Tilde{\theta}_{n, t}\right)$}
                \State Send $\Delta \check{\theta}_{n, t}$ to the PS
            \EndFor
            \State \underline{\textbf{Server Side (PS):}}
            \For{$n=1,...,N$}
                \State Receive $\Delta{\check{\theta}_{n, t}}$ from client $n$
            \EndFor
            \State Global Averaging: $\Delta\check{\theta}_{t+1} = \frac{1}{N} \cdot \sum_{n=1}^{N} \Delta{\check{\theta}_{n, t}}$
            \State Send $\Delta\check{\theta}_{t+1}$ to clients
        \EndFor
    \end{algorithmic}
\end{algorithm}

\subsection{Extending \textit{FedMap} for Heterogeneous Data Distributions}
\label{subsec:fedmap_feddr}
Given recent developments in addressing challenges associated with learning  from non-IID data, we combine our \textit{FedMap} method with the \textit{FedDR}\cite{feddr} method to handle statistical heterogeneity. 
By using the Douglas-Rachford splitting technique, FedDR aims to solve the common composite objective of the original FL optimisation problem, which involves the addition of some regularizer $\mathcal{L}(\theta) + g(\theta)$, where the authors propose $L1$ regularisation. 
Across our experimental evaluation, we find that $L1$ regularisation led to sub-optimal results, hence we omit the composite setting. 
Additionally, we do not employ the randomised block-coordinate technique proposed in \cite{feddr}, since this would require client sub-sampling which is beyond the scope of this work.

Without applying $g(\theta)$, applying FedDR involves the following three steps:

\textit{1) Intermediate variable update}: After the client reconstructs the global model, $\Tilde{\theta}_t$, each client updates an intermediate variable, $\Tilde{\theta}^y_{n, t}$, influenced by both the global model, and the client's prior model $\Tilde{\theta}_{n, t-1}^y$. The update is given as:

\begin{equation}
    \Tilde{\theta}_{n, t}^y := M_t \cdot \Tilde{\theta}_{n,t-1}^y + \alpha (\Tilde{\theta}_{t} - M_t \cdot \Tilde{\theta}_{n, t-1})
    \label{eq:feddr_variable}
\end{equation}
accumulating information from both the local and global models. Initially, $\Tilde{\theta}_{n, 0}^y=\Tilde{\theta}_{t=1}$, the global model at initialisation. 
The gradual update helps address the non-convexity of the problem by preventing drastic swings in model parameters. 
The step size, $\alpha$, determines how quickly new knowledge is incorporated.
We incorporate updating of the intermediate variable in \textit{FedMap} after line 8 in Algorithm \ref{alg:fedmap}.

\textit{2) Proximal operator application}: 
A quadratic penalty is added to each clients' loss for each batch, forming the proximal operator. 
This is then applied to the intermediate variable to obtain local model updates; augmenting the local loss function:
\begin{equation}
    \begin{aligned}
        \Tilde{\theta}_{n, t} & := \text{prox}_{\eta f_c}(\Tilde{\theta}_{n, t}^y) \\
         & := \underset{\theta}{\text{arg min}} \quad \mathcal{L}_n(\Tilde{\theta}_{n, t}) + \frac{1}{2\eta} \| \Tilde{\theta}_{n, t} - \Tilde{\theta}_{n,t}^y \|^2
    \end{aligned}
    \label{eq:feddr_update}
\end{equation}

The 
quadratic penalty term 
is inversely modulated by $\eta$ where smaller values for $\eta$ align the local model more with $\Tilde{\theta}^y_{n, t}$, balancing local adaptivity and global coherence.

\textit{3) Model reflection}: To promote exploration, $\Tilde{\theta}_{n, t}$  and $\Tilde{\theta}^y_{n, t}$ are combined in the reflection step:

\begin{equation} \Tilde{\theta}^x_{n, t} = 2 \cdot \Tilde{\theta}_{n, t} - \Tilde{\theta}^y_{n, t} \label{eq:feddr_realignment}
\end{equation}

Finally, the difference $\Delta \Tilde{\theta}^x_{n, t} = \Tilde{\theta}^x_{n, t} - M_t \cdot \Tilde{\theta}^x_{n, t-1}$ , is sent back to the server for aggregation, where $\Delta \Tilde{\theta}^x_{n, t}$ replaces $\Delta \Tilde{\theta}_{n, t}$ in line 10 in Algorithm \ref{alg:fedmap}.

\section{Experiments}
\label{sec:experiments}
The experiments are organized into four categories: examining the impact of 1) pruning schedules and 2) FedMap's hyperparameters, 3) evaluating FedMap's performance in a non-IID deployment scenario, and 4) benchmarking against \emph{FederatedPruning} \cite{FederatedPruning}.

\subsection{Exp. A: Step-wise and Continuous Pruning Schedules}
\label{subsec:pruning_schedules}
\begin{figure}[t]
    \centerline{\includegraphics[width=0.75\columnwidth]{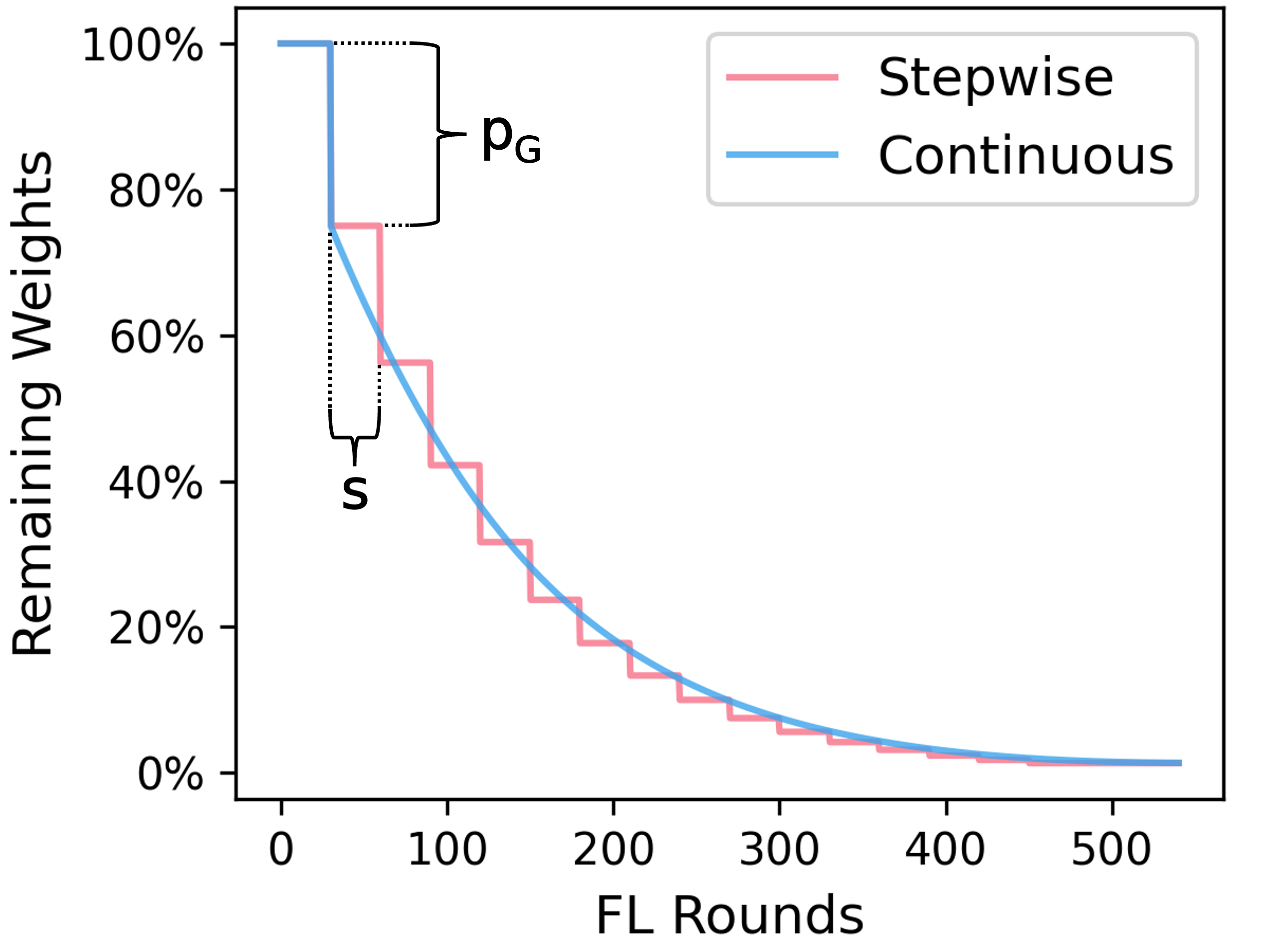}}
    \caption{The stepwise and continuous pruning schedules exemplified. We design the schedules, s.t. the model can be trained to a good initial solution at the start. After $\mathbf{s}$ FL rounds, clients prune the global model further, s.t. $(1-p_g)\cdot d$ parameters remain. We unanimously use an exponential series, where $p_g = 0.25$, progressively removing 25\% of the remaining model parameters.}
    \label{fig:pruning_schedules}
\end{figure}

In our first experiment we study the impact of two different pruning schedules under different pruning cadences (varying $\mathbf{s}$, see Fig. \ref{fig:pruning_schedules}) on global model performance.
Central to the experiment was an exploration of two distinct pruning schedules. The first schedule, known as the step-wise schedule, involves a stepping function that determines when the model should be pruned after a certain number of FL rounds. The step width of the model, denoted as $\mathbf{s}$, follows the procedure outlined in Algorithm \ref{alg:imp}.

Secondly, we adopt a continuous pruning schedule, offering a smoother and more gradual pruning of the model parameters. 
We utilize Bézier interpolation to approximate the step-wise pruning schedule over the $T$ FL-iterations in our experiments. 
With this continuous schedule, we closely match the cumulative impact of the pruning process over $T$, effectively reducing the model at a comparable pace, ensuring that at any given time, the model retains a similar level of unpruned parameters. 
This smooth schedule aims for a more uniform transition in pruning intensity.

In this first experiment, we trained 10 clients on IID splits of the CIFAR-10 dataset \cite{cifar10}. 
Each client participates in a set of $L=4$ epochs of local training at every round $t$, before contributing to the global model update. 
Furthermore, we compared the influence of pruning schedules on ResNet-56 \cite{resnet} and MobileNetV2 \cite{mobilenet_v2} network architectures\footnote{ https://github.com/chenyaofo/pytorch-cifar-models}.
The models were trained using Stochastic Gradient Descent (SGD) with a learning rate of $\eta = 0.01$, weight decay rate of $5e^{-4}$ and batch size of $128$. 

We experimented with the following pruning cadences; $\mathbf{s} = \{25, 30, 35, 45, 90\}$, to assess the impact on global model convergence in terms of overall performance and maximally achievable pruning ratios of the models. 
Additionally, we fix the pruning proportion to $p_G = 0.25$, which means that over the step width $s$, an additional 25\% of the model parameters are pruned with respect to the unpruned parameters.
We also took precautions to set a defined pruning limit. 
Precisely, models were pruned up to a maximum of 1\% (100x compression) of their total parameters accepting potential performance implications.

\subsection{Exp. B: Influence of local training epochs and pruning cadence}
\label{subsec:training_e_and_pruning_schedules}
Our second experiment aimed to explore the impact of modulating $L$ alongside different pruning cadences. 
Specifically, we tested for $L=\{2,4,8,16\}$ while simultaneously adjusting $s=\{45,90,135\}$, using the CIFAR-10 dataset as well as IID splits of the CIFAR-100 \cite{cifar10} and SVHN \cite{SVHN} datasets to expand our empirical scope. 
To reduce the number of FL iterations, we re-calibrated the pruning magnitude of the models to reach 5\%, since beyond this point the drop in performance is catastrophic.  
Based on the outcomes of the first experiment, we favoured the step-wise pruning methodology over its continuous counterpart. 
Finally, to make the setup more representative, we increased the number of clients from 10 to 30.

\subsection{Exp. C: Non-IID Experiments}

In this set of experiments, we examine the performance of FedMap on two common non-IID (Non-Independently and Identically Distributed) benchmarks, which remark a more realistic scenario within real-world deployments.
For this, we use two datasets i.e., \emph{i)} the FEMNIST (Federated Extended MNIST), and \emph{ii)} the Shakespeare dataset \cite{LEAF}.
FEMNIST is a modified version of the EMNIST \cite{emnist} dataset. It expands the $10$ class digit problem of the MNIST \cite{mnist} dataset by encompassing $62$ classes, including uppercase and lowercase characters. 
We follow the partitioning scheme proposed in \cite{LEAF}, which involves grouping characters and digits by their respective author.
In the context of Federated Learning (FL), each author's data corresponds to the data on an FL client. 
This embodies a real-world scenario in which data is inherently partitioned between different devices.

Our second dataset is the Shakespeare dataset, which includes all of William Shakespeare's works. 
Similarly to the FEMNIST dataset, the Shakespeare dataset is organized based on a logical entity: the speaking roles in the plays \cite{LEAF}. 
Each speaking role is assigned to a different FL client, which simulates a situation where each client device has data of a distinct categorical nature.
Table \ref{tab:non_iid_datasets_statistics} provides a comprehensive summary of the dataset statistics. 
It is imperative to note that the ``Number of Slices" in the table corresponds to the character writers in the FEMNIST dataset and individual speakers in the Shakespeare dataset. 
This representation aids in understanding the distribution and structure of the datasets in a manner conducive to FL experimentation. 
The values for mean and standard deviation suggest that FEMNIST presents a more uniformly distributed dataset across clients, offering a benchmark for assessing our methodologies across varying degrees of non-IID data.

\begin{table}[t]
\centering
\caption{Statistics of datasets in LEAF \cite{LEAF}}
\label{tab:non_iid_datasets_statistics}

\begin{tabular}{lcccc}
\toprule
Datasets & \# Slices & Total Slices & \multicolumn{2}{c}{Samples/Slice} \\
\cmidrule(lr){4-5}
& & &  Mean & Std. \\
\midrule
FEMNIST & 3,550 & 805,263 & 226.83 & 88.94 \\
Shakespeare & 1,129 & 4,226,158 & 3,743.28 & 6,212.26 \\
\bottomrule
\end{tabular}
\end{table}

We subsample $150$ slices from either of the datasets. Due to the significant volume of unused data, we employ a 60/40 train/test split, which translates to 90 FL clients for training and 60 evaluation slices for testing. 
Performance assessment is carried out by averaging the results across all evaluation-reserved slices, providing a robust evaluation of all FL configurations. 

\textbf{Models}: For the FEMNIST dataset, we use the proposed model in LEAF \cite{LEAF}, with two convolutional layers followed by two fully connected layers. 
For the Shakespeare dataset, characters are mapped to an eight-dimensional embedding, then processed by a two-layer Long Short-Term Memory network (LSTM) \cite{LSTM} with a sequence length of $80$. 
The LSTM outputs are mapped back to the vocabulary for next-character prediction. 
Unlike in the original implementation, we added batch-normalisation \cite{batch_norm} for added stability and faster convergence, as well as dropout \cite{dropout} for regularisation to our setup.

\textbf{Hyperparameters}: For all experiments, data is batched with a size of $10$ to align with the settings proposed in \cite{feddr} and we use Stochastic Gradient Descent (SGD) for optimisation. 
A learning rate of $3e^{-4}$ and $3e^{-3}$ is used for the two datasets, respectively, and we set the dropout probability for the LSTM with $0.5$. 
For the FEMNIST experiments we chose $L=20$ to align with FedDR, and based on LEAF \cite{LEAF} we opt for $L=1$ for the Shakespeare dataset.

The performance of models trained on the Shakespeare and FEMNIST datasets responded significantly to changes in the hyperparameters for FedDR and FedMap (see Section \ref{subsec:fedmap_feddr}). 
We observe that smaller step sizes such as the ones used in experiment A, B, and C were too aggressive to maintain performance and was further exacerbated for the FEMNIST dataset. To ensure sufficient convergence and comparison, the step size has been titrated for both datasets such that the models are able to learn enough before the first pruning round to prevent an immediate and unrecoverable drop in performance. For Shakespeare this was $\mathbf{s} = 135$ and $\mathbf{s} = 270$ for FEMNIST.

As described previously, the two main hyperparameters for FedDR are $\eta$ and $\alpha$. 
We select these based on a small grid search around the parameters of the original implementation for FEMNNIST in \cite{feddr}, that is, $\eta=1000, \alpha = 0.95$.
Surprisingly, the original parameters were largely unstable in our setup for FEMNIST, but more suitable for Shakespeare. 
FEMNIST benefitted from a much lower $\eta$ at $10$, while Shakespeare performed best at the proposed $\eta=1000$. 

For $\alpha$ a value of $1$ and $1.2$ was found to be the best for Shakespeare and FEMNIST respectively. Increasing $\alpha$ for FEMNNIST leads to higher convergence but far less stability, while there is no significant difference with Shakespeare.

After a certain number of pruning iterations, subsequent rounds in which pruning takes place are markedly characterized by a drop in performance and decreased stability. In FEMNIST, this occurs at the first pruning step, for Shakespeare this is the second. We hypothesize that the performance of the parameters chosen for FedDR may become unfaithful at these rounds and may prove to be detrimental. Therefore, we propose several hybrid methods to alter the training dynamics at these rounds:
\begin{enumerate}
    \item Switching from FedDR to FedAvg.
    \item Modifying the values of $\eta$ and $\alpha$.
\end{enumerate}

For models trained on the Shakespeare dataset, adjusting $\alpha$ had a minor impact on convergence, whereas varying $\eta$ had a much larger impact. 
Here smaller values for $\eta$ seemed beneficial, leading to better stability, faster convergence and better maximal performance. 
Performance improved as $\eta$ increased. Specifically, the best results were obtained when we completely removed the effect of the proximal operator and therefore regularisation; just leaving the alterations in training dynamics via the model reflection step.
In contrast, FEMNIST experienced minimal effects from the alteration of regularization (via adjustments $\eta$, and the best results were found without change, i.e. $\eta=10$. 
Furthermore, increasing $\alpha$ surprisingly promoted stability, with the best value being $1.75$. 

As a concluding extension, we investigated weighting local updates by each client's dataset size, aiming to equalise the influence of clients on the model, regardless of their data volume.
Specifically, this extension is applied to Shakespeare, given its significantly larger sample skew; aiming to prevent overfitting to predominant speaking-roles.

All of the above configurations are summarised in Table \ref{tab:feddr_configs}. 
Additionally, since the proximal term (see Eq. \ref{eq:feddr_update}) represents the $L2$ difference between the previous local model and the global model, this is skewed if the model is pruned in the respective round. 
This is because the pruned model has fewer trainable parameters. 
We took this into account and masked the parameters of the local model of the previous round with the same pruning mask, preventing any differences in pruned weights to accumulate. However, there was no effect on training.

\newcolumntype{C}{>{\centering\arraybackslash} X }
\begin{table}[t]
\centering
\caption{Specific FedDR configurations. Dynamic configurations change at the first pruning iteration for FEMNIST, and the second pruning iteration for Shakespeare.}
\begin{tabularx}{\linewidth}{lX}
\toprule
\multicolumn{1}{l}{\textbf{Name}} & \multicolumn{1}{l}{\textbf{Configuration Details}} \\
\midrule
FedDR & Basic FedDR implementation, no pruning. \\
\midrule
FedMap-FedDR & Basic FedDR with FedMap pruning. \\
\midrule
FedMap-FedDR-C1 & FedDR with pruning, changes to FedAvg. \\
\midrule
FedMap-FedDR-C2 & FedDR with pruning, changes $\alpha$ and $\eta$ to optimal values. \\
\midrule
FedMap-FedDR-C3 & FedMap-FedDR-C2, with updates weighted by local dataset size. \\
\bottomrule
\end{tabularx}
\label{tab:feddr_configs}
\end{table}

\begin{figure*}[htp]
  \centering
  \begin{minipage}{\textwidth}
    \begin{subfigure}{0.33\textwidth}
      \includegraphics[width=\textwidth]{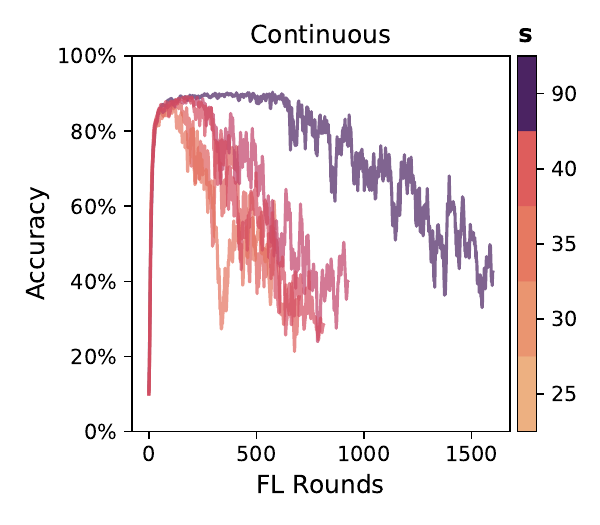}
    \end{subfigure}%
    \hfill
    \begin{subfigure}{0.33\textwidth}
      \includegraphics[width=\textwidth]{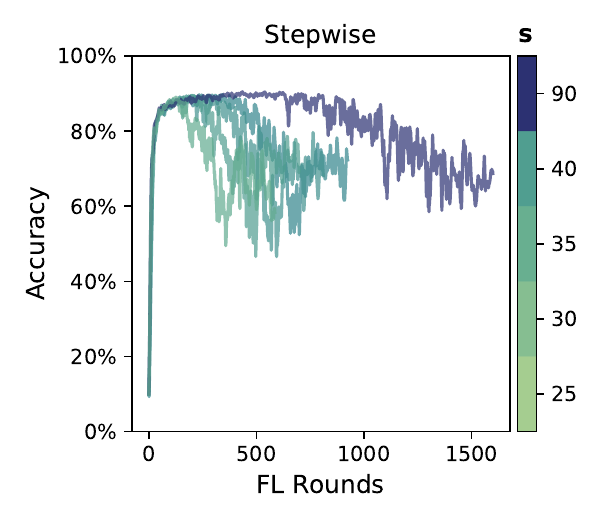}
    \end{subfigure}%
    \hfill
    \begin{subfigure}{0.33\textwidth}
      \includegraphics[width=\textwidth]{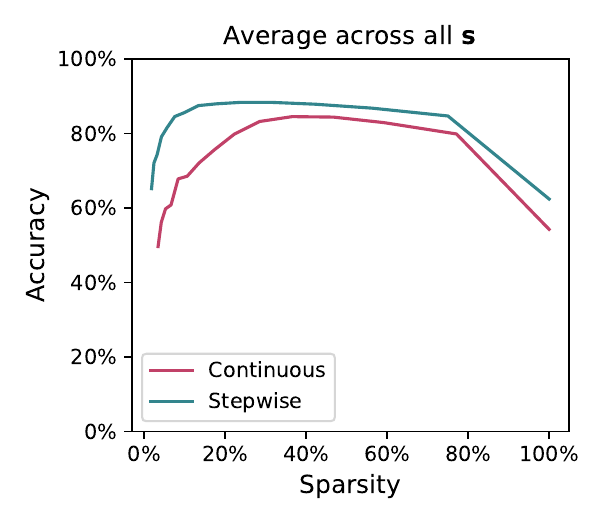}
    \end{subfigure}
  \end{minipage}
  \vspace{-1em}  
  \begin{minipage}{\textwidth}
    \begin{subfigure}{0.33\textwidth}
      \includegraphics[width=\textwidth]{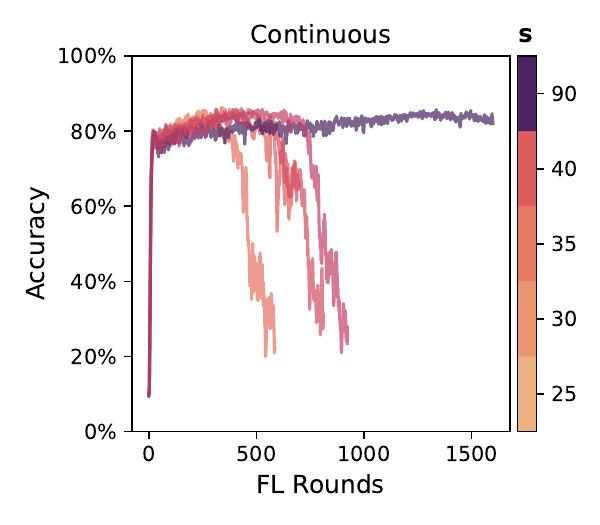}
    \end{subfigure}%
    \hfill
    \begin{subfigure}{0.33\textwidth}
      \includegraphics[width=\textwidth]{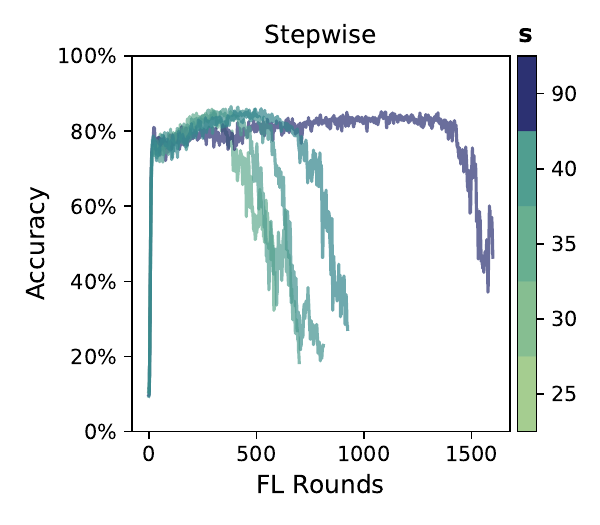}
    \end{subfigure}%
    \hfill
    \begin{subfigure}{0.33\textwidth}
      \includegraphics[width=\textwidth]{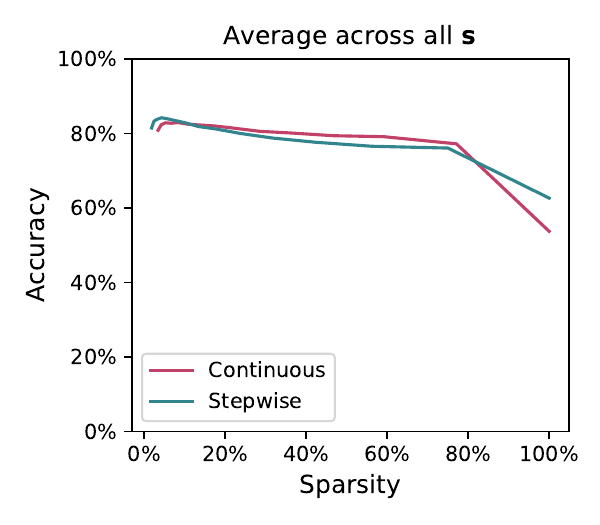}
    \end{subfigure}
  \end{minipage}
  \caption{Comparison of pruning schedules trained on CIFAR10. Top row is with Resnet56 and the bottom row is with MobileNet.}
  \label{fig:initial_experiment}
\end{figure*}

\subsection{Exp. D: Benchmarking against FederatedPruning}
We also performed experiments comparing against the \textit{FederatedPruning} method proposed by Lin et al. \cite{FederatedPruning}. 
The key differences between \textit{FederatedPruning}  and \textit{FedMap} are as follows:
\begin{enumerate}
    \item \textit{FederatedPruning} suggests pre-training and inital pruning on a centralised server before  beginning the FL procedure, akin to what \cite{PruneFL} propose, which may require settings where data exists sufficiently.
    \item \textit{FederatedPruning} prunes the global model after global aggregation on the PS, whereas in \textit{FedMap}, pruning happens on the clients, eliminating the need to communicate pruning masks.
    \item \textit{FederatedPruning} proposes a novel structured ranking method to determine the in- or exclusion from the set of prunables according to the $p_G$ in every round; whereas in \textit{FedMap} we rely on the state-of-the-art in unstructured pruning, namely the \textit{LAMP} method proposed in \cite{LAMP}.  
    \item \textit{FederatedPruning} allows for the re-activation of previously pruned locations, whereas in \textit{FedMap} all pruning masks from progressive pruning are subsets of preceding pruning masks.
\end{enumerate}

To allow for a fair comparison, we operate \textit{FederatedPruning} with \textit{LAMP} instead of their proposed structured pruning methodology, which we denote as \textit{FederatedPruning-UnstructuredLamp} in our experiments.
Additionally, we chose their structured pruning method targeting half-columns, which we denote as \textit{FederatedPruning-StructuredColumnHalf}, and which the authors chose within their experimental settings. 

We conducted two experiments, for two different models. 
First, we train the MobileNetV2 model using the SVHN dataset, representing a relatively simple task with an architecture optimised for mobile deployments \cite{mobilenet_v2}.
This combination of a lightweight model and a less complex dataset emulates scenarios that may arise in real-world mobile deployments. 
Secondly, we train the ResNet56 model on CIFAR100 dataset, combining a larger model architecture with increased task-complexity. 
We follow the same pruning schedules and training-parameters as outlined in our previous experiments.

\begin{figure*}[t]
    \centering
    \begin{minipage}{.49\textwidth}
        \centering
        \textbf{Resnet56}
    \end{minipage}
    \begin{minipage}{.49\textwidth}
        \centering
        \textbf{MobilenetV2}
    \end{minipage}

    \begin{subfigure}[b]{\textwidth}
        \centering
      \includegraphics[width=.49\textwidth]{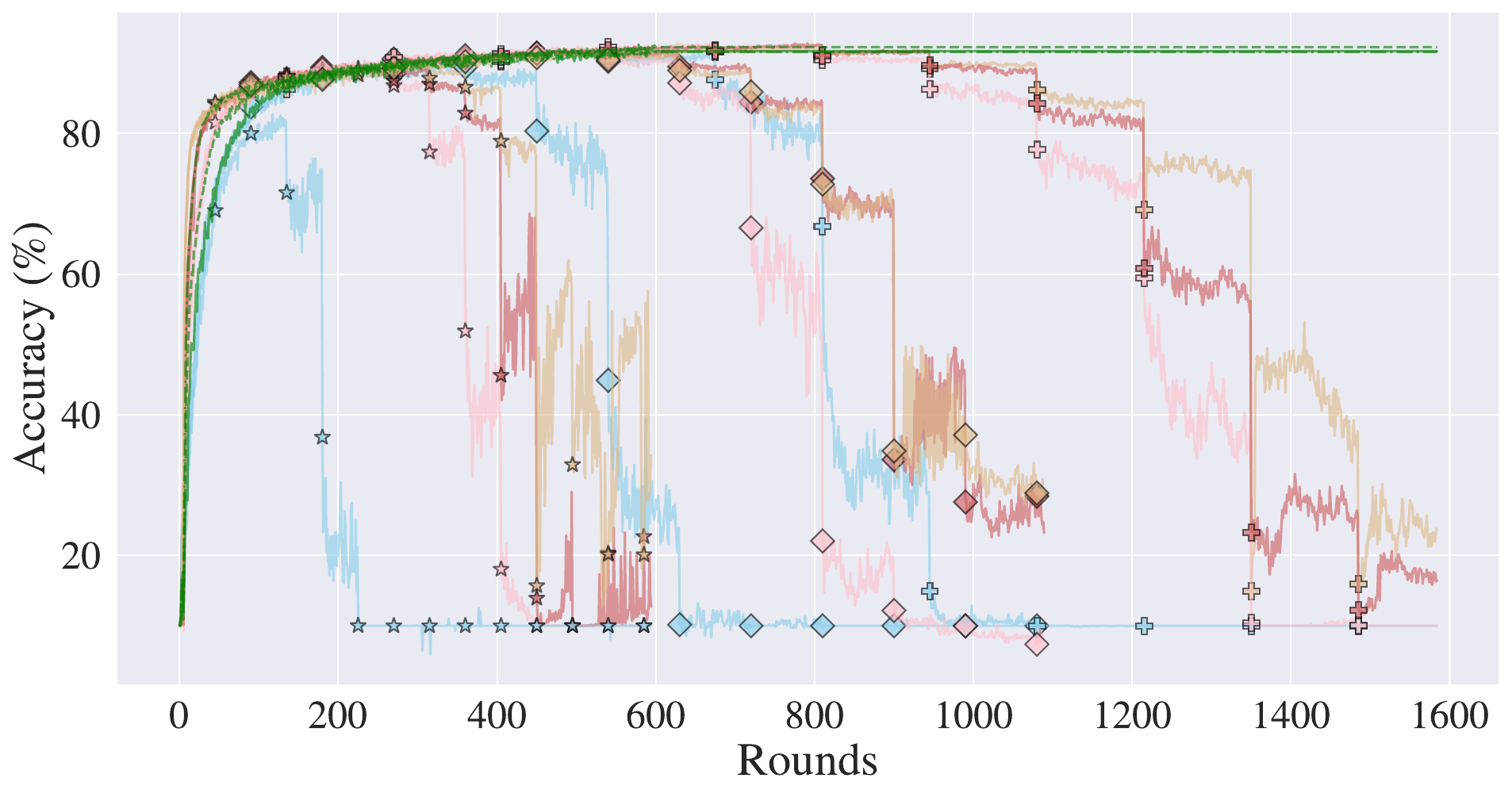}
      \includegraphics[width=.49\textwidth]{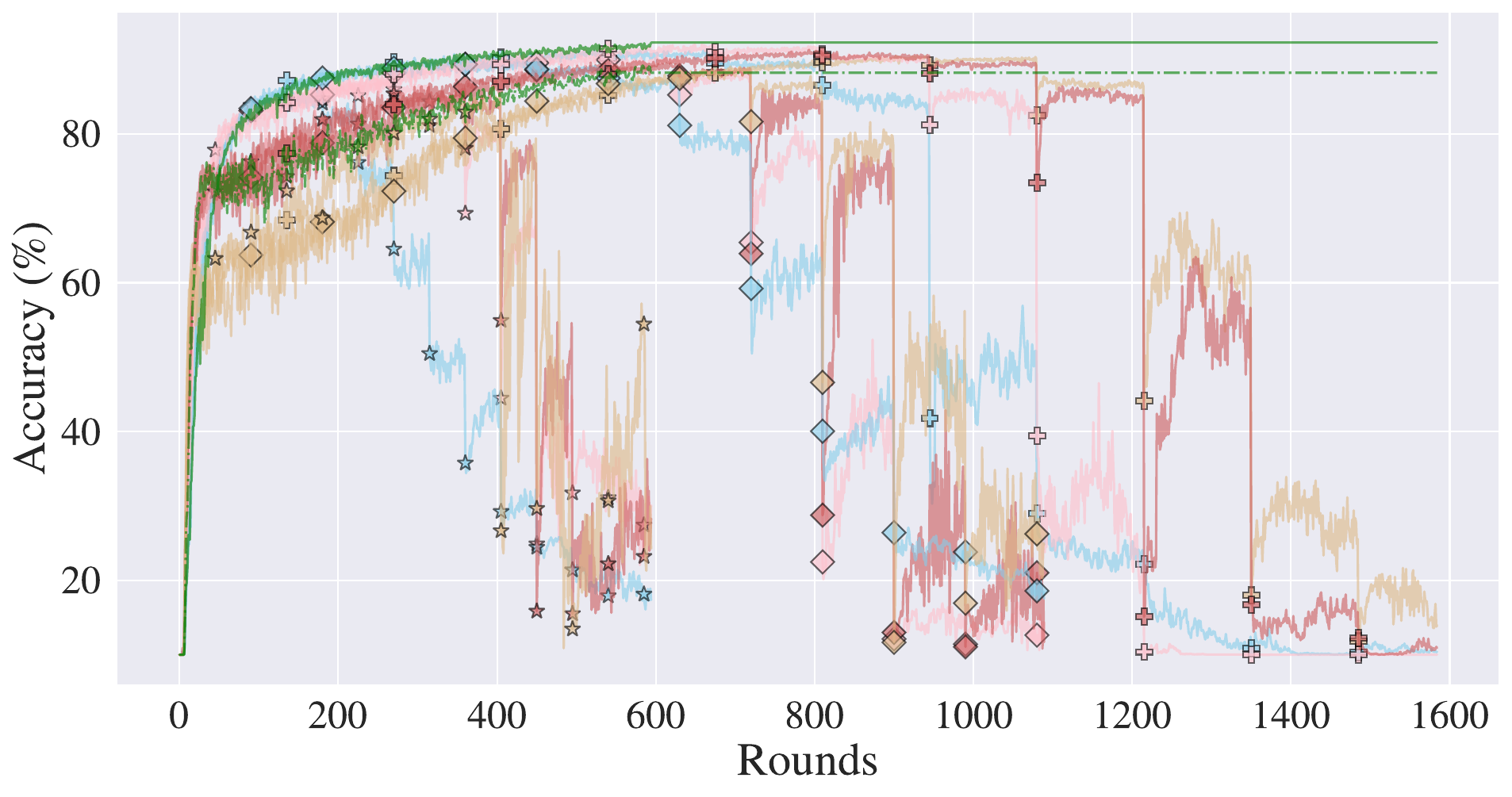}
      \label{fig:fedmap_cifar10_acc_rnds}
    \end{subfigure}

    \includegraphics[width=\textwidth, trim={0 7cm 0 7cm},clip]
    {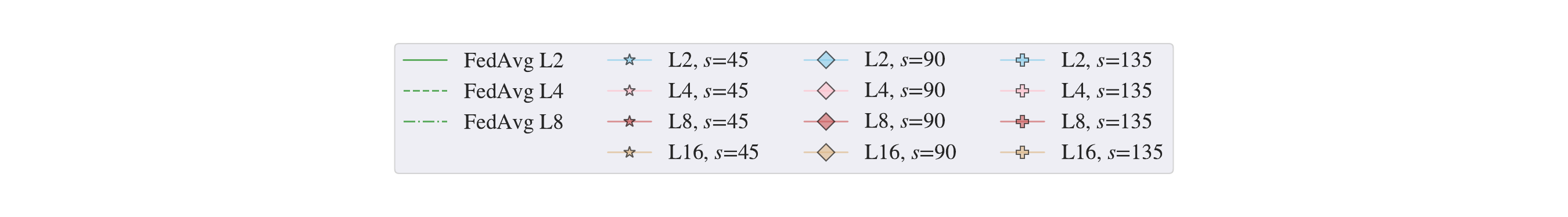}

  \caption{Accuracy vs. rounds across variable number of local epochs ($L$) and step-widths ($s$) for CIFAR10.}
  \label{fig:fedmap_main_results}
\end{figure*}

\section{Results and Discussion}
\label{sec:results_disc}
\subsection{Investigating the Impact of Pruning Schedules on Model Performance and Stability}
In our first experiment, we assessed the influence of the type of pruning schedule on the model's ability to retain performance while progressing in pruning. 
To this end, we conducted experiments using two image classification models: Resnet56 \cite{resnet} and MobileNetV2 \cite{mobilenet_v2}. 
Our findings were consistent across both models, revealing some key insights. 
Expectedly, increased pruning induced a noticeable decline in model performance. 
We discovered better overall performance by extending the intervals between pruning iterations (larger $\mathbf{s}$). 
This positive effect was observed even at higher pruning rates, implying that less frequent pruning leads to better performance at even higher pruning rates.

The two models produced varying results when comparing model sparsity and accuracy. 
On average (across all interval sizes $\textbf{s}$), the step-wise pruning schedule outperformed the continuous schedule. 
Leading to faster convergence and similar performance (see MobileNetV2, bottom row in Figure \ref{fig:initial_experiment}) or increased average performance (see ResNet56, top row in Figure \ref{fig:initial_experiment}). 
This trend is especially evident at high pruning rates ($p_G \leq 20\%$).

A critical observation was made that both models maintained their performance effectively up to a sparsity level of 80\%, meaning that they retained only 20\% of their original parameters. 
After crossing a certain threshold, we observed a significant reduction in the model's performance. 
Despite this, we proceeded with pruning beyond the 80\% sparsity level, allowing us to gain insights into how varying the step-width $\mathbf{s}$ affects the model's performance under extreme sparsity conditions, and to understand better the impact of different pruning schedules on the model's performance.
After discovering that larger $\mathbf{s}$ allow for more substantial pruning (retaining higher performance longer), we investigate the underlying reasons for this effect. 

Less frequent but more substantial pruning can improve model performance by providing a stable learning and adaptation environment. 
When pruning occurs less frequently, the models have more time to adjust and optimize their remaining parameters in response to the reduced network complexity. 
This uninterrupted learning phase is crucial for allowing the models to more effectively assimilate the information from the training data, leading to a more robust learning process \cite{optim_pruning_in_nns}. 
As a result of reducing the frequency of pruning, the final performance of the models can be improved. 
Frequent pruning can cause disruption and hinder performance optimization within each pruned state. 
By allowing models to settle and optimize their performance within each state, they can maintain high accuracy levels, allowing for higher pruning rates.

\begin{figure*}[t!]
   
    \centering
    
    \includegraphics[width=1\textwidth]
    {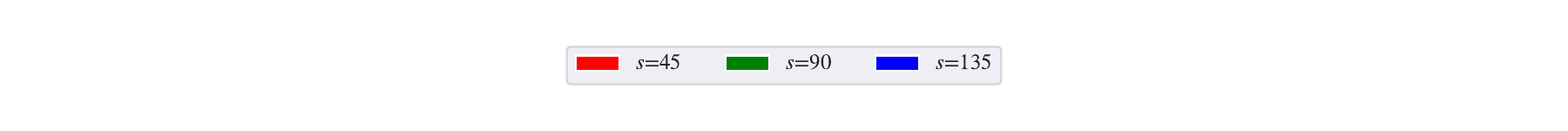}
    
    \begin{subfigure}[b]{\textwidth}
        \centering
      \includegraphics[width=\textwidth, trim={0 3em 0 0},clip]{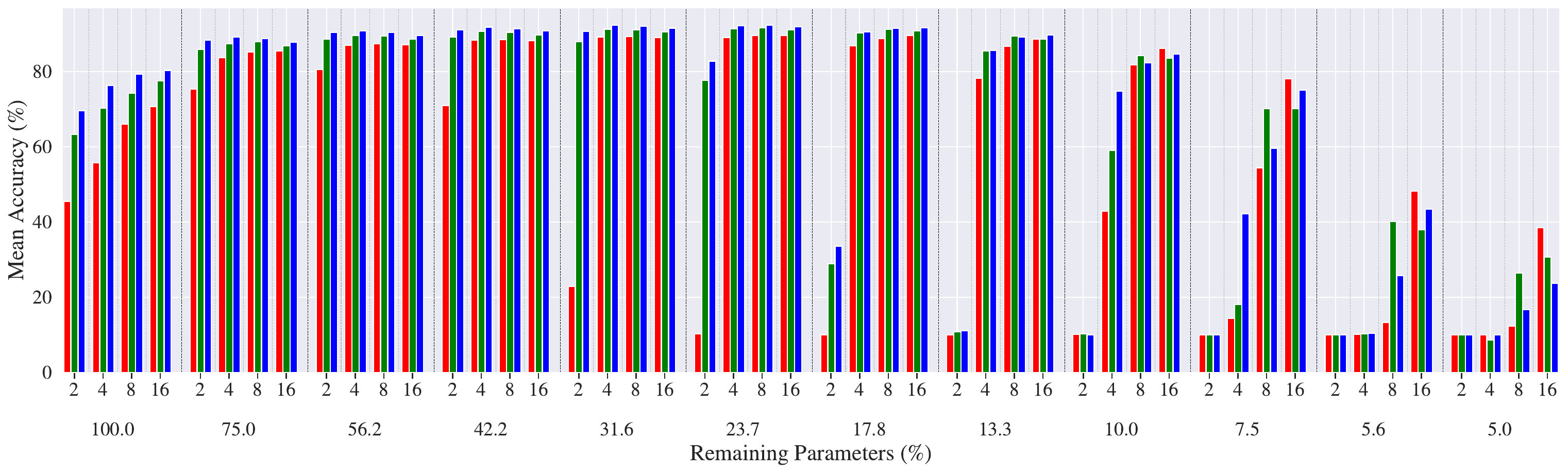}
      \vspace*{-8.5ex}  
        \begin{flushleft}
        \tiny
        \hspace{1.5em}$\mathbf{L}$\\
         \vspace{-0.5em}
        \dotfill \\
        Remaining \\
        Parameters \\
        \hspace{1.5em}(\%)
        \end{flushleft}
       \caption{Resnet56}
    \end{subfigure}

     \vspace{1em}
    
        \begin{subfigure}[b]{\textwidth}
    \setcounter{subfigure}{3}
        \centering
      \includegraphics[width=\textwidth, trim={0 3em 0 0},clip]{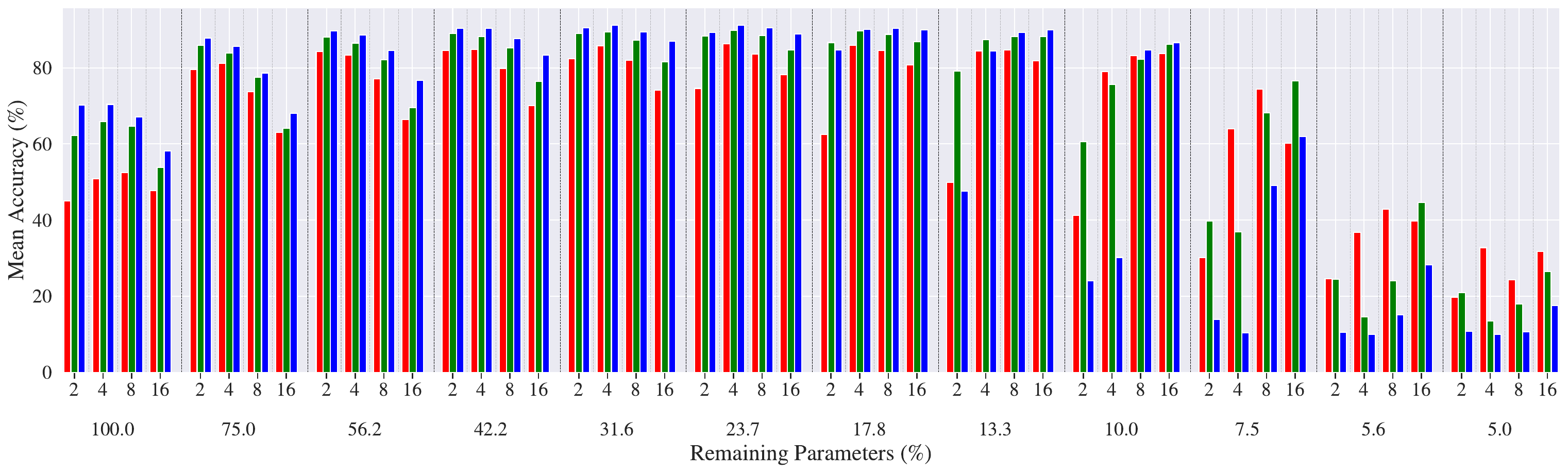}
      \vspace*{-8.5ex}  
        \begin{flushleft}
        \tiny
        \hspace{1.5em}$\mathbf{L}$\\
        \dotfill \\
        Remaining \\
        Parameters \\
        \hspace{1.5em}(\%)
        \end{flushleft}
       \vspace{-0.5em}
         \caption{MobileNetV2}

    \end{subfigure}
    
    \caption{Average performance per pruning step ($s$) across variable different values of local epochs ($L$) for CIFAR10.}
    \label{fig:cifar10_bars}
\end{figure*}

\begin{figure*}[h!]
        \centering
        \begin{minipage}{0.5\textwidth}
            \centering
            \textbf{Resnet56}
        \end{minipage}%
        \begin{minipage}{0.5\textwidth}
            \centering
            \textbf{Mobilenetv2\_x0\_5}
        \end{minipage}
    
        \vspace{0.5em}
        
        \begin{subfigure}[b]{\textwidth}
            \centering
          \includegraphics[width=.49\textwidth]{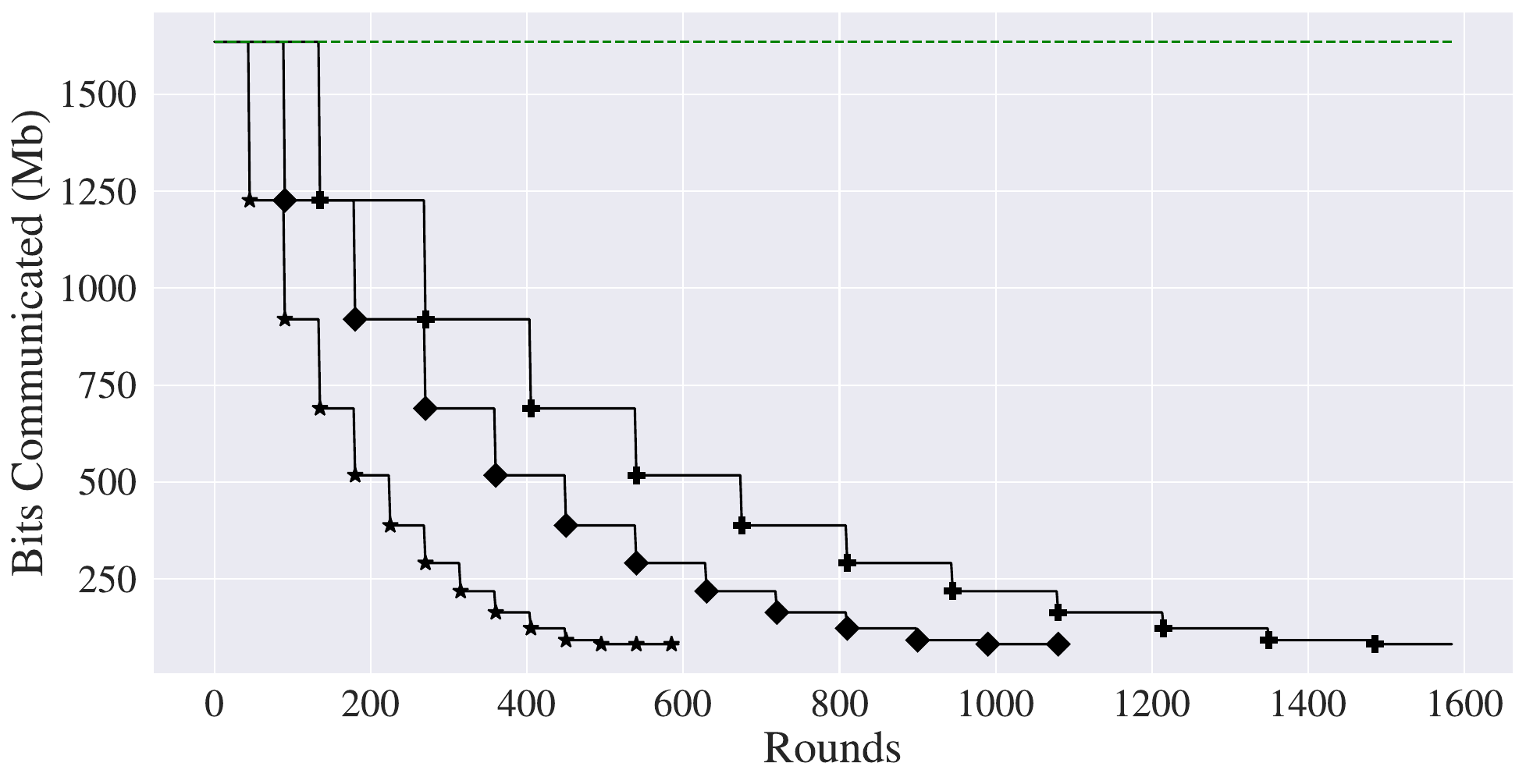}
          \includegraphics[width=.49\textwidth]{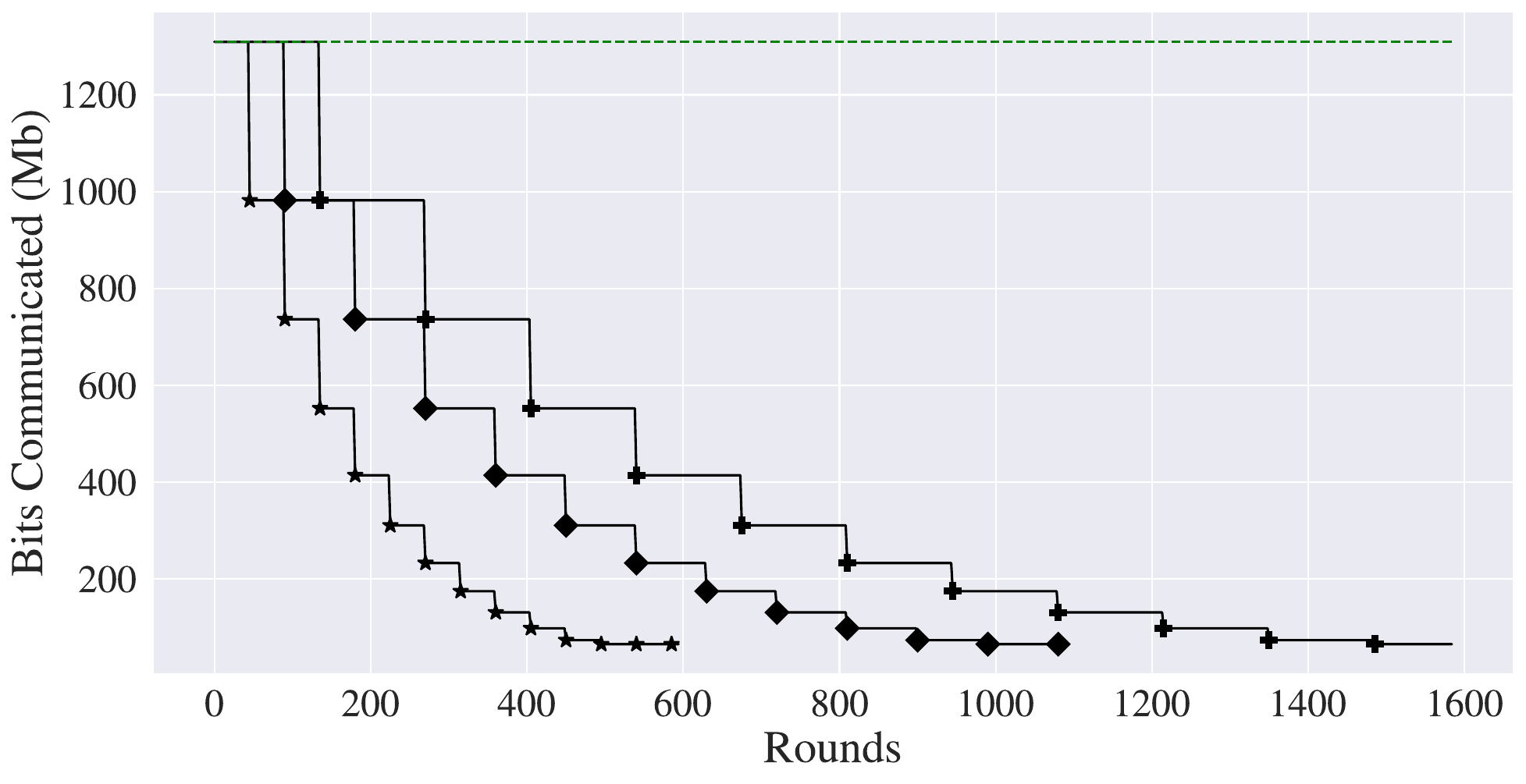}
          \vspace{-0.3em}
        \end{subfigure}    

    \includegraphics[width=1\textwidth]{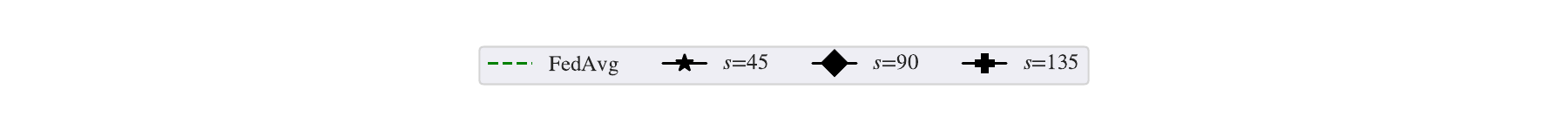}
    \caption{Communicated bits (Mb) per round of FL for exp. B. on CIFAR10.}
    \label{fig:fedmap_main_comms}
\end{figure*}

\subsection{Balancing Pruning Frequency and Local Training Regimes}
Figure \ref{fig:fedmap_main_results} (combined with Figure \ref{fig:fedmap_main_results_appendix} in the appendix) illustrate our findings indicating a significant relationship between task complexity, convergence, and performance retention under high pruning rates. According to our analysis, models trained on the 
CIFAR100 dataset may experience early performance degradation (see Figure \ref{fig:fedmap_cifar100_acc_rnds}). 
However, models trained on the CIFAR-10 and SVHN datasets achieve peak performance more quickly and maintain superior performance for a longer period (see Figures \ref{fig:fedmap_main_results_appendix} and \ref{fig:fedmap_svhn_acc_rnds}). 
Models trained on the latter can even withstand substantial pruning without significant loss of performance. 
This phenomenon highlights the importance of model resilience in the face of pruning, indicating considerable variation across different tasks.

Furthermore, the number of local training epochs emerges as a pivotal factor; increasing the number of local training epochs, $\mathbf{L}$, enhances the model's ability to preserve high performance over time. 
However, prolonged local training can increase the demand for client resources, such as energy and compute, or increase the risk of model divergence. 
Excessive training on local datasets can cause local models to deviate from the global model, potentially reducing the overall performance across client devices and leading to a degradation of the global model performance. 
While strategies to mitigate this issue of model divergence have been proposed, such as those in \cite{Ma2021FederatedLW} and \cite{Xu2021RobustMA}, they were not the focus of this work. 
Careful tuning of the local training epochs $\mathbf{L}$ is therefore crucial to strike a balance between preserving model performance under pruning and minimizing the associated resource demands and risks of model divergence.

Our research has found that there is a certain point at which model performance experiences a drastic decline due to excessive pruning. 
Before this threshold is reached, the models can withstand pruning without significant performance loss; however, after reaching this critical point, model performance is irreversibly impaired. 
While the issue of performance degradation following pruning is well-known in model pruning (see \cite{LAMP}), deploying our FedMap methodology requires the inclusion of fallback mechanisms to mitigate performance loss and ensure consistent delivery of expected performance levels. 
This critical challenge must be addressed when deploying \textit{FedMap} in real-world scenarios.

In Figure \ref{fig:cifar10_bars}, we demonstrate the delicate balance between pruning ratios, pruning intervals $\mathbf{s}$ and local training epochs $\mathbf{L}$ on the performance of Resnet56 and MobileNetV2 models trained on CIFAR10. 
Generally, a larger $\mathbf{s}$ leads to better convergence at the beginning and to models retaining higher overall performance for up to $p_G \geq 20\%$. 
Beyond this point, and especially for pruning ratios where more than 90\% of the parameters are pruned, smaller $\mathbf{s} \in \{45, 90\}$ lead to retaining better performance. 
This phenomenon suggests that prolonged training may unintentionally lead to over-fitting and, in turn, diminish the model's effectiveness at high pruning levels.

Additionally, we also performed the same experiment on the SVHN and CIFAR100 datasets (see Figures \ref{fig:svhn_cifar100_resnet_bars} and \ref{fig:svhn_cifar100_mobilenet_bars} in the appendix). 
For instance, a ResNet56 model trained on the SVHN task can withstand pruning down to 7.5\% of its original parameter count with minimal performance loss. 
Similarly, a MobileNetV2 model can be aggressively pruned to 5\% of its original size (20x compression), accepting an 8.5\% decrease in performance. 
When evaluating models designed for the CIFAR100 dataset, it is evident that pruning beyond 75\%  leads to severe performance degradation. 
This threshold highlights a crucial point at which reducing model parameters becomes counterproductive. 
Furthermore, our investigation reveals that extending the number of FL iterations (larger $T$) before pruning ($\mathbf{s} = 135$) offers marginal benefits in performance; for higher pruning ratios, a shorter pruning interval ($\mathbf{s}$) proves more advantageous, maintaining better performance under such rigid compression conditions. 

Moreover, when we increased the local training epochs ($\mathbf{L}$) for a model, it became better equipped to handle higher pruning rates. 
This highlights the importance of on-device training, as it helps to improve the model's ability to withstand significant parameter reductions. 
On the contrary, larger pruning intervals $\mathbf{s}$ may lead to overall higher top-performance. 
When considering very large pruning ratios, smaller $\mathbf{s}$, lead to higher performance at higher pruning ratios. 
Furthermore, we can see that the complexity of the task at hand also affects how tolerant a model is to pruning.

\begin{figure*}[t]
        \begin{subfigure}[b]{.49\textwidth}
            \centering
          \includegraphics[width=\linewidth]{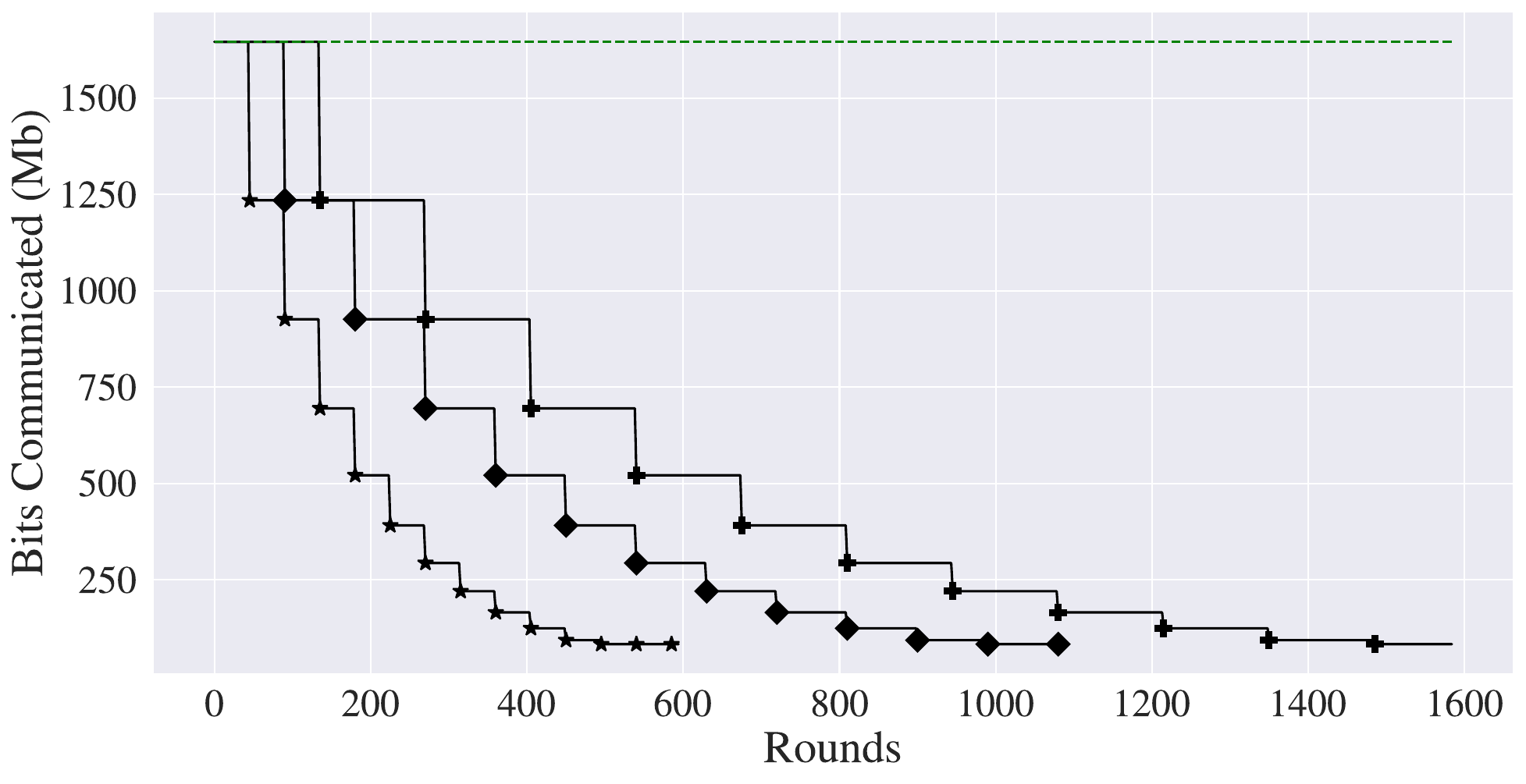}
          \caption{CIFAR100}
        \end{subfigure}
        \begin{subfigure}[b]{.49\textwidth}
            \centering
          \includegraphics[width=\linewidth]{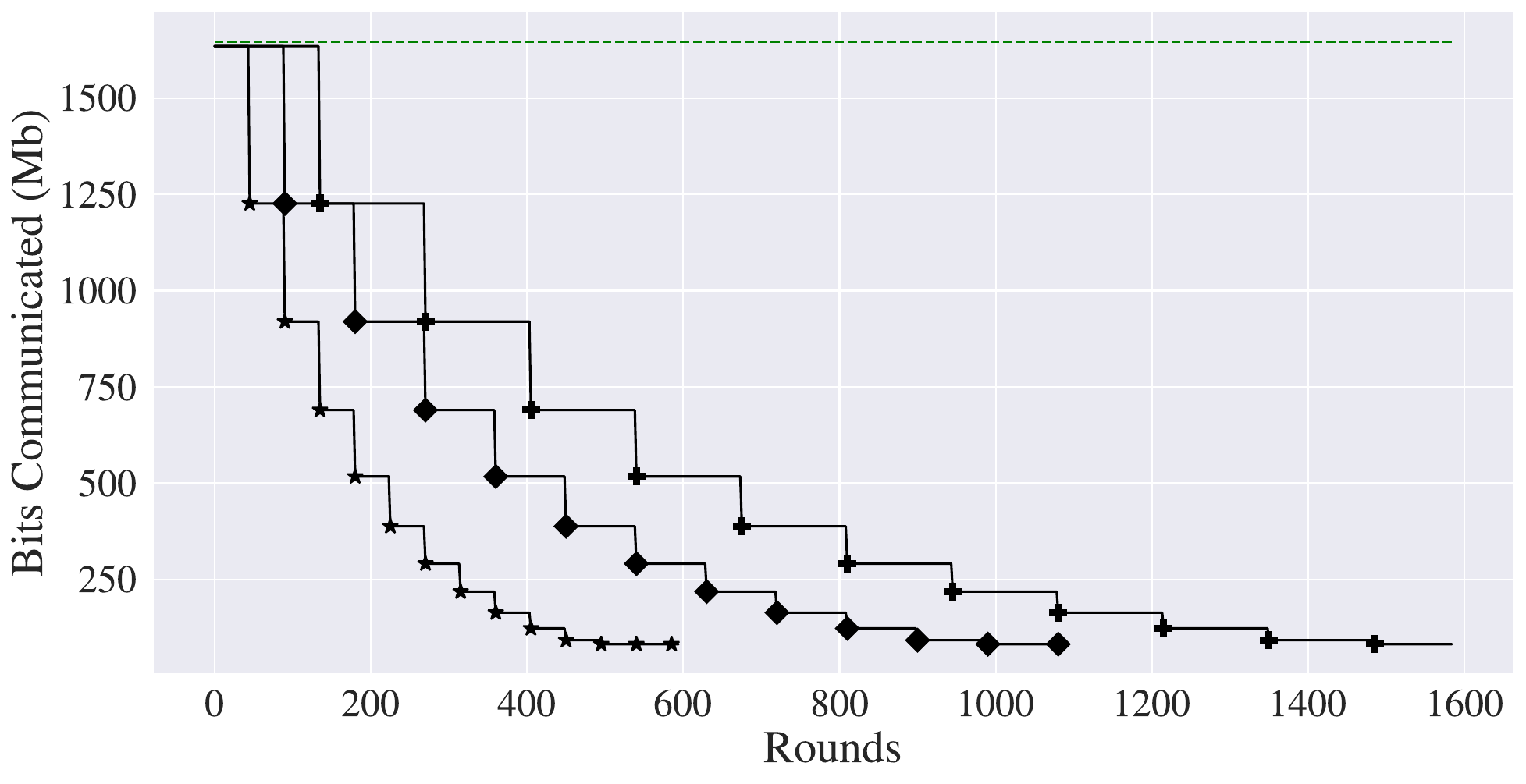}
          \caption{SVHN}
        \end{subfigure}

     \includegraphics[width=1\textwidth]{figs/experiment_2/comms_legend.pdf}
    
    \caption{Communicated bits (Mb) per round of FL for exp. B. for Resnet56.}
    \label{fig:fedmap_main_comms_cifar100}
\end{figure*}

In Figure \ref{fig:fedmap_main_comms}, we present a quantitative analysis of the bytes communicated for varying values of the step width, $\mathbf{s}$. 
Note, the CIFAR10 and SVHN datasets are both problems with 10 classes each, hence the number of parameters is equivalent (for results on the CIFAR100 and SVHN datasets, see Figure \ref{fig:fedmap_main_comms_cifar100}). 
A lower value for $\mathbf{s}$ leads to faster pruning and less data transmitted. 
This efficiency underscores the importance of optimising $\mathbf{s}$ to reduce communication overhead in FL settings.
Consistent with our prior findings, increasing the number of local training epochs (denoted by a large $\mathbf{L}$) is instrumental. 
Extended local training allows clients to refine the model more thoroughly before aggregation, enhancing the model's resilience to pruning-induced performance decline.
However, it is imperative to recognise the potential drawbacks of excessively elongating the local training phase. 
Overly extensive local training periods can inadvertently impair the global model's performance, a phenomenon well-documented in existing literature \cite{Zhang2021AdaptiveFL, Yan2022OnDeviceLF, Wu2022GlobalUG, Ma2021FederatedLW, Xu2021RobustMA}. 
This delicate balance emphasises the need for strategic selection of $\mathbf{L}$, ensuring it is sufficient to support effective pruning while avoiding the counterproductive effects of over-training on the model's aggregate performance.

\begin{figure*}[t]
    \centering
    \begin{minipage}{0.5\textwidth}
        \centering
        \textbf{Shakespeare}
    \end{minipage}%
    \begin{minipage}{0.5\textwidth}
        \centering
        \textbf{FEMNIST}
    \end{minipage}

    \vspace{0.5em}
  
    \begin{subfigure}[b]{\textwidth}
        \centering
      \includegraphics[width=.495\textwidth]{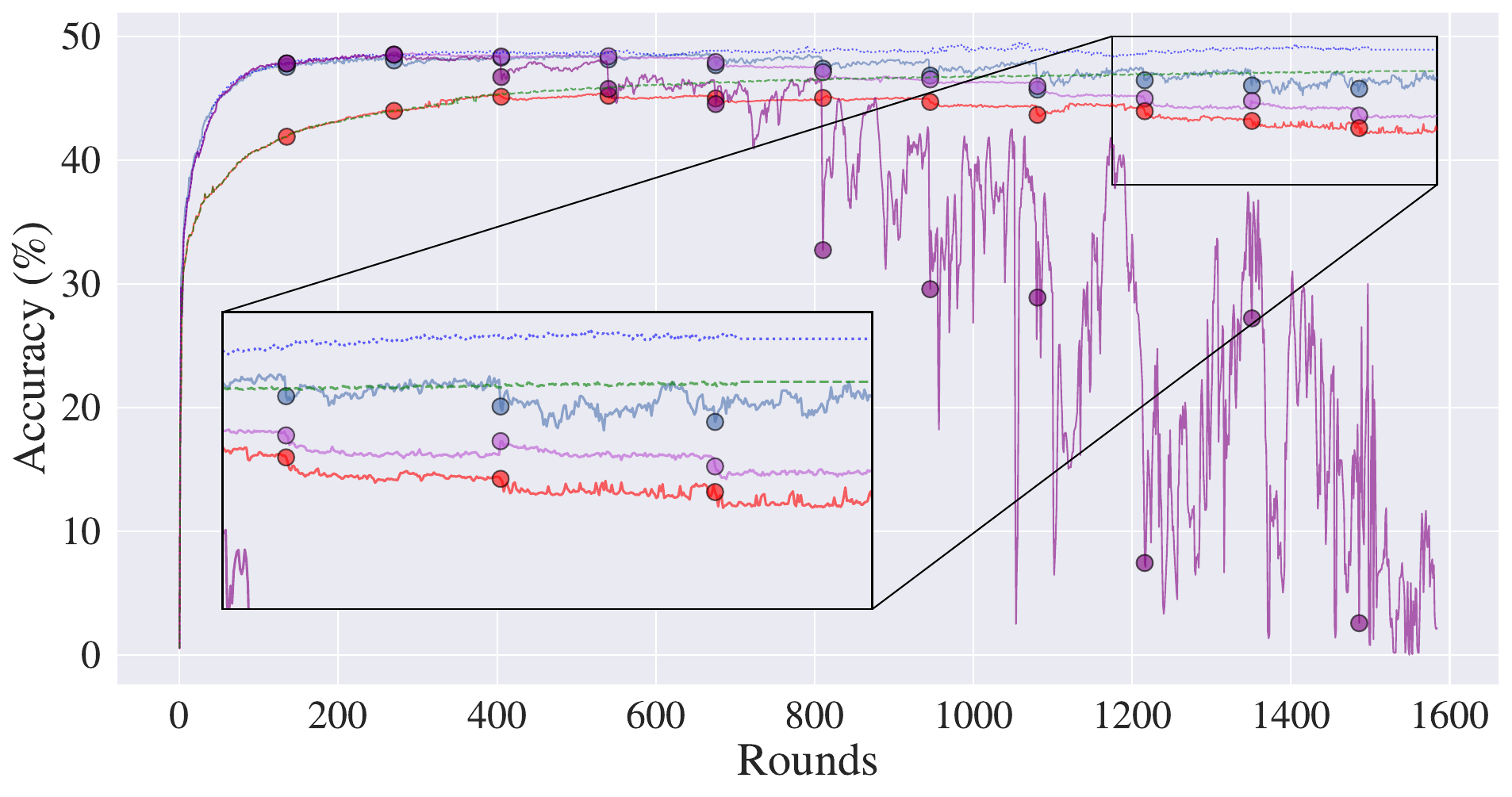}
      \includegraphics[width=.495\textwidth]{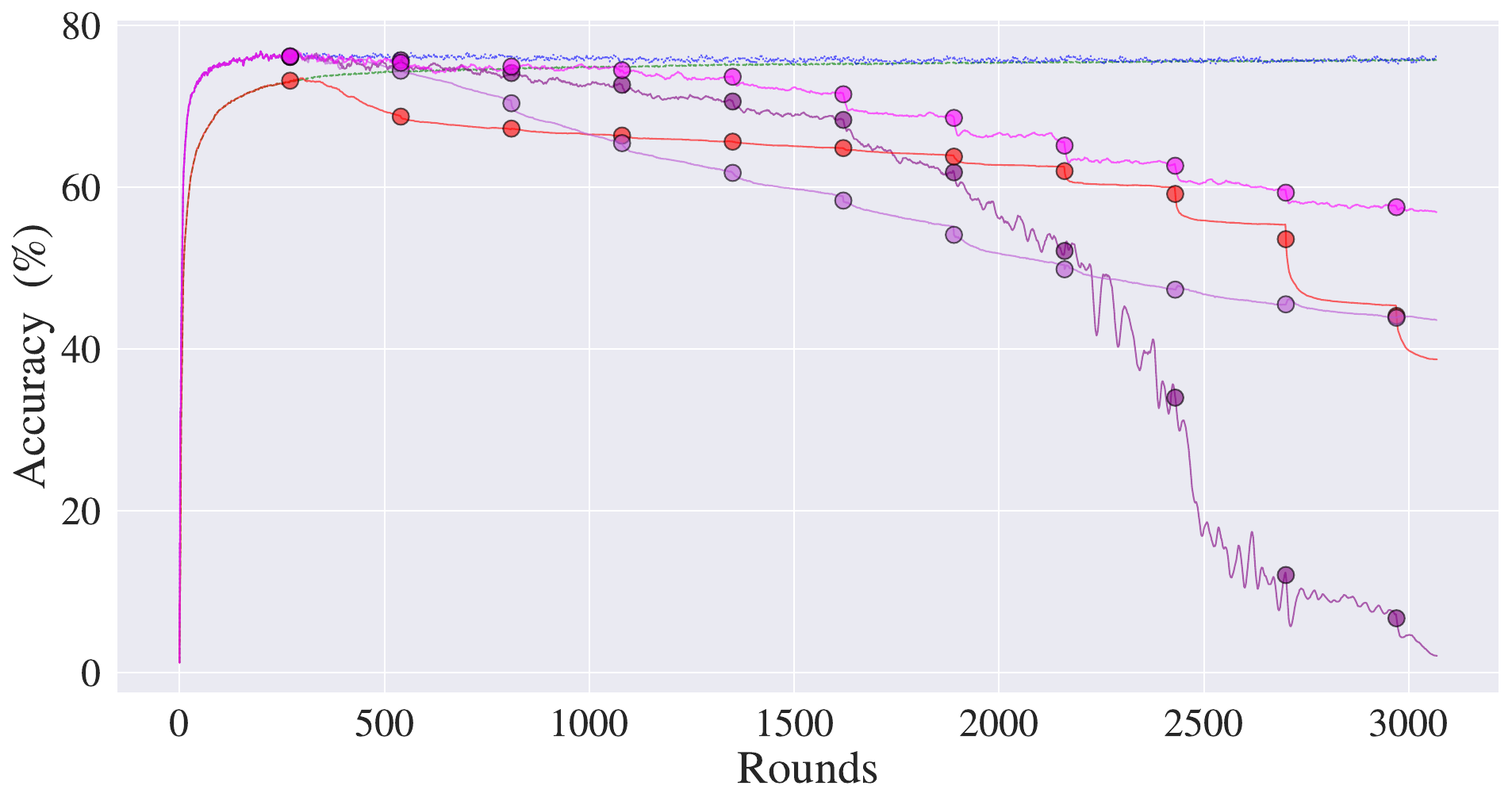}
      \vspace{-0.5em}

    \end{subfigure}%

    \includegraphics[width=\textwidth]
    {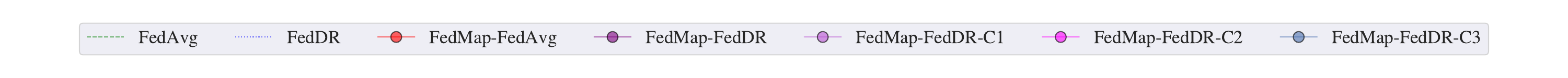}
  \caption{Accuracy vs. rounds for the Non-IID experiments. Refer to table \ref{tab:feddr_configs} for FedDR configurations.}
  \label{fig:non_iid_experiments}
\end{figure*}

\subsection{Integrating FedDR into FedMap for Enhanced Non-IID Pruning}
Figure \ref{fig:non_iid_experiments} depicts our non-IID results for Shakespeare (left) and FEMNIST (right) datasets. 
For both, as previously mentioned, the basic \textit{FedMap-FedDR} approach sharply drops in accuracy after the first few pruning points. 
Configuration \textit{FedDR-C1}, that switches to FedAvg after these pruning points, was able to significantly stabilise at later pruning rounds and outperformed \textit{FedMap-FedDR} for Shakespeare. 
For FEMNIST, while it was able to stabilise until the subsequent pruning point, it then experienced more rapid decline, more than \textit{FedMap-FedDR}. 
However, this decline was more consistent and the final result was marginally higher than \textit{FedMap-FedDR}. 

Our approach to further adapting the hyperparameters, in configurations \textit{FedMap-FedDR-C2} -- FEMNIST, and \textit{FedMap-FedDR-C3} -- Shakespeare, resulted in more desirable results. 
For both datasets, these methods were able to provide large improvements in stability and preserving significantly more accuracy than \textit{FedMap-FedDR}. 
For the Shakespeare dataset using \textit{FedMap-FedDR-C3}, we were able to achieve similar accuracy to FedAvg, showcasing the potential synergy between FedMap and FedDR.
For FEMNIST, we achieve at least 80\% pruning (after 5 pruning steps i.e., by gradually removing 25\% of the weights in the previous step).

Although these are promising results for preserving accuracy in the context of pruning, the optimal hyperparameter and performance seem highly dataset dependent, as evidenced by their response to different settings. 
Currently, it is very challenging to achieve a generalisable setup for maintaining performance, but our experiments suggest numerous opportunities for further tailoring these methods in the future.

\begin{figure*}[t]
  \centering 
    \begin{minipage}{0.45\textwidth}
        \centering
        \textbf{MobileNetV2, SVHN}
    \end{minipage}
    \begin{minipage}{0.46\textwidth}
        \centering
        \textbf{Resnet56, CIFAR100}
    \end{minipage}

    \vspace{0.5em}
  
    \begin{subfigure}[b]{\textwidth}
        \centering
      \includegraphics[width=.495\textwidth]{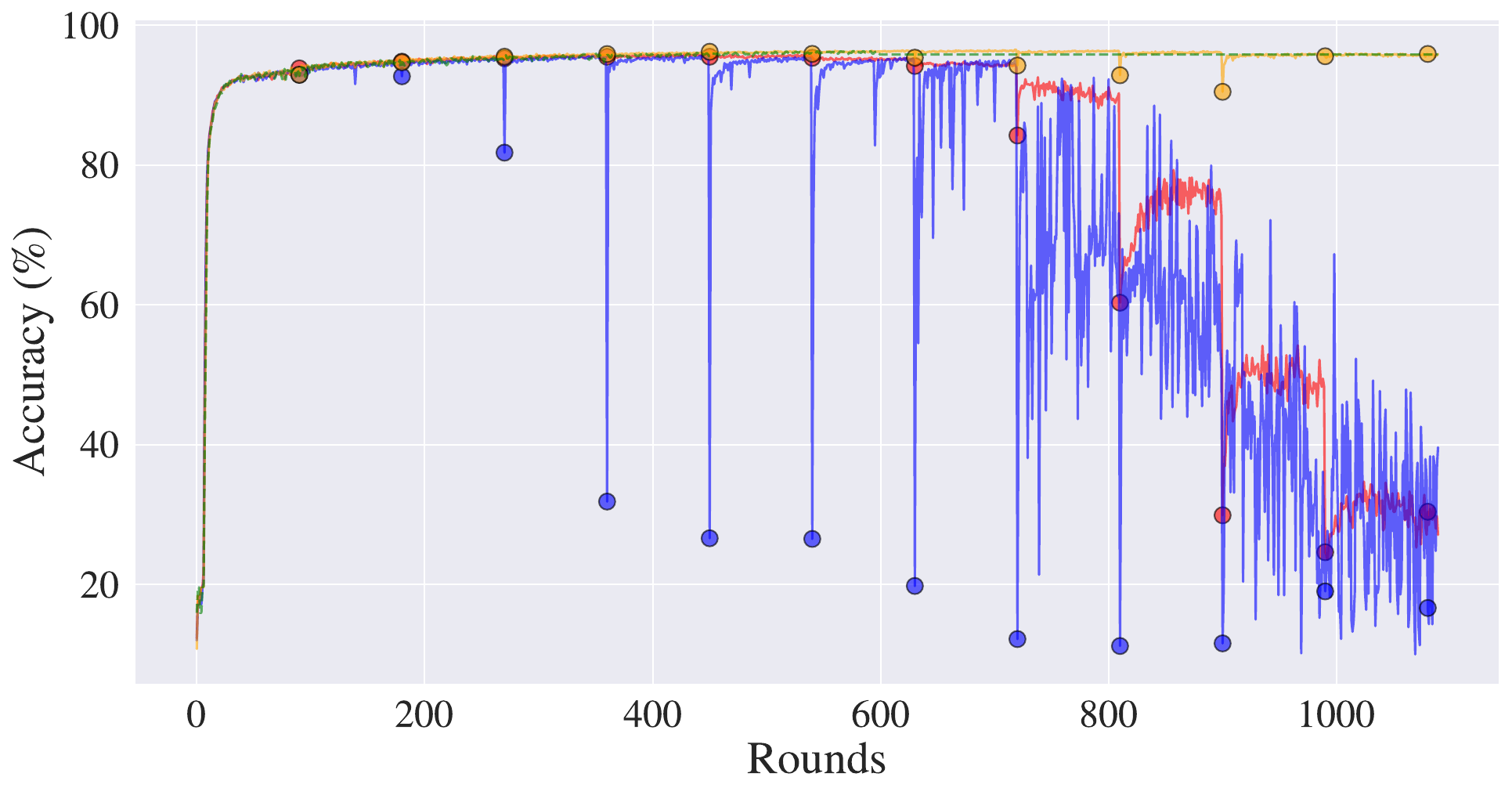}
      \includegraphics[width=.495\textwidth]{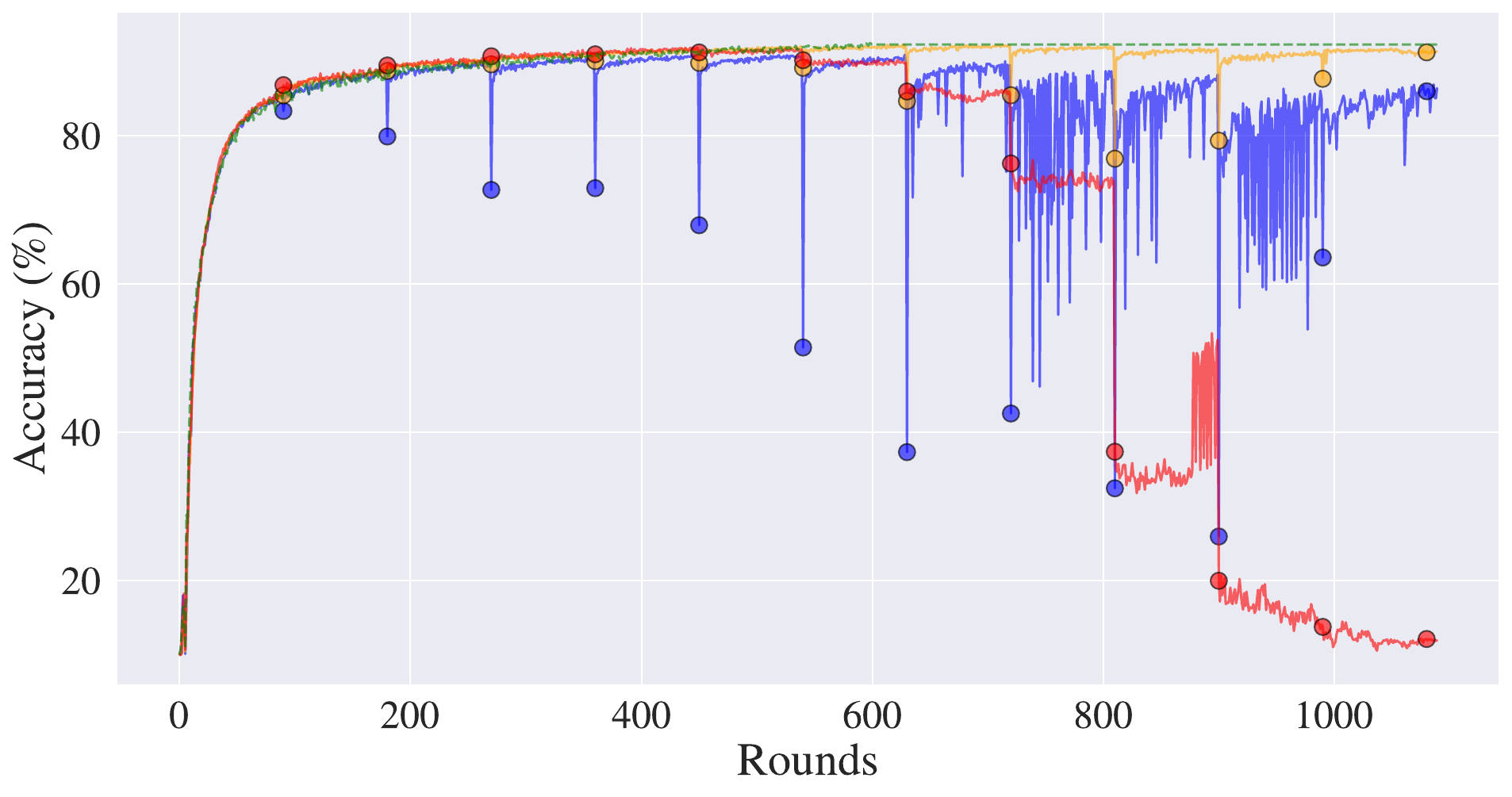}
      \vspace{-0.5em}
      \caption{Accuracy vs. Rounds}
    \end{subfigure}

    \includegraphics[width=1\textwidth]
    {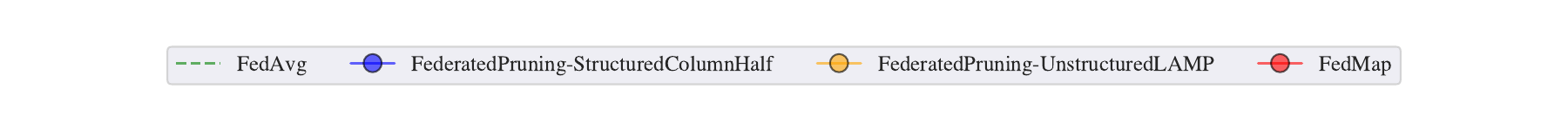}
    
    \vspace{1em}
    
    \begin{subfigure}[b]{\textwidth}
        \centering
      
      \includegraphics[width=.495\textwidth]{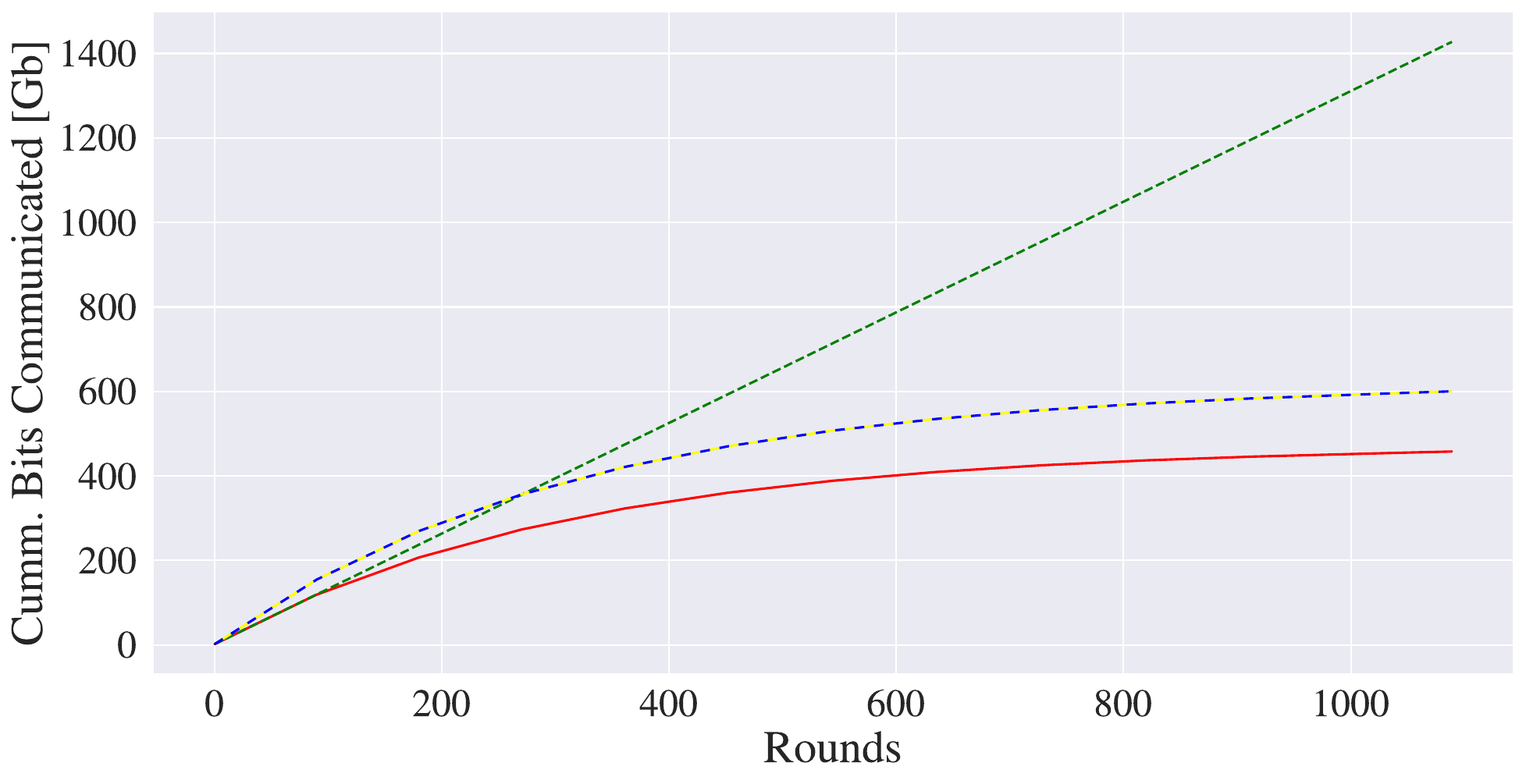}
      \includegraphics[width=.495\textwidth]{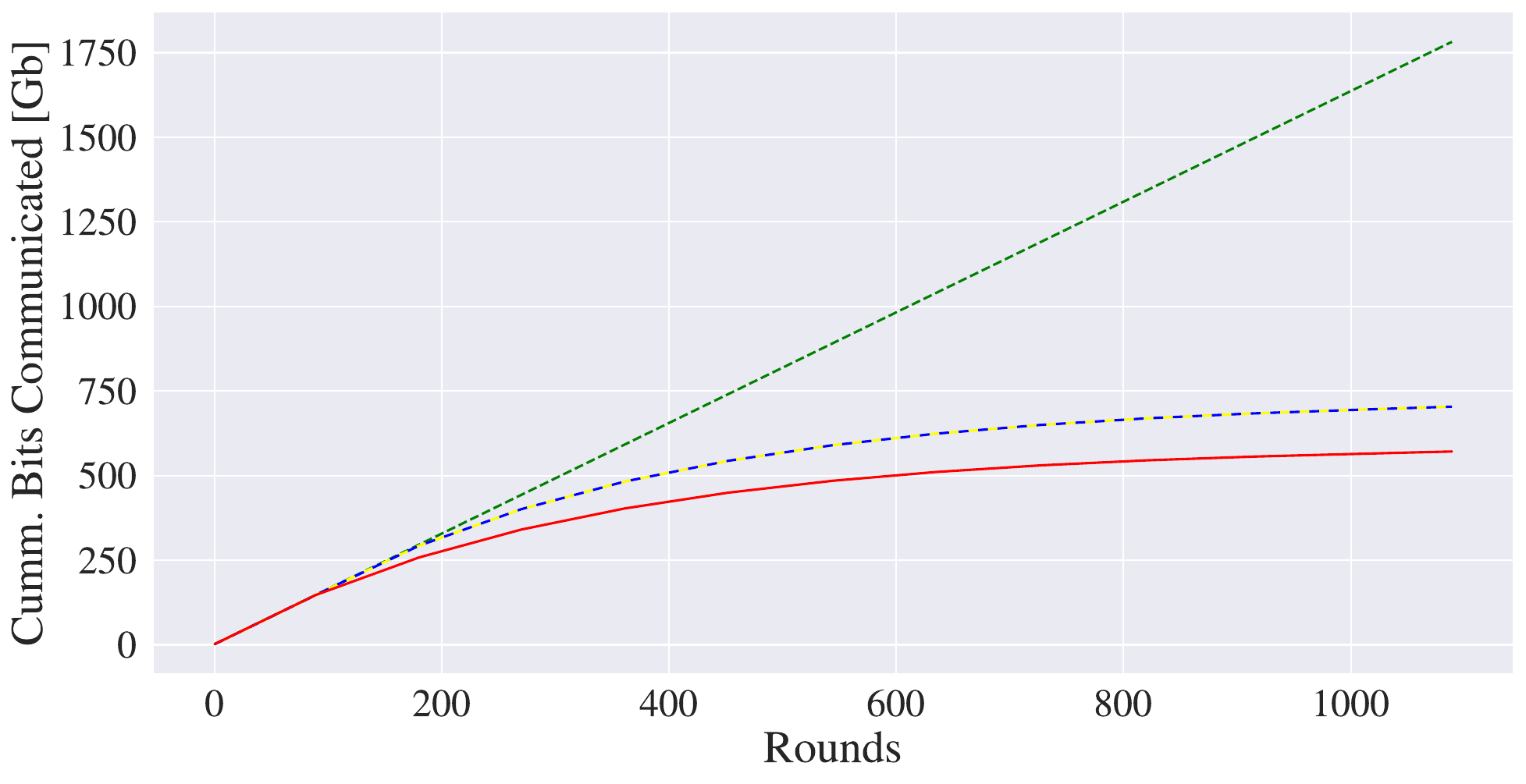}
      \vspace{-0.5em}
      \caption{Total bits communicated (Mb) vs. Rounds}
    \end{subfigure}
    
    \includegraphics[width=1\textwidth]
    {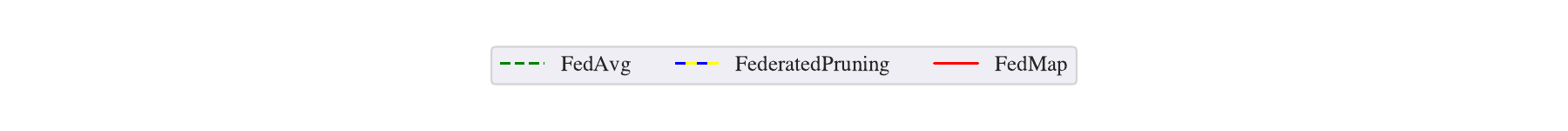}
    \vspace{-2em}
    
  \caption{Comparison of accuracy and communication cost for Federated Pruning benchmark (Exp. D).}
  \label{fig:federated_pruning_experiments}
\end{figure*}

\subsection{Evaluating FedMap Against Unstructured FederatedPruning}
To ensure appropriate evaluation, we modified the \textit{FederatedPruning} \cite{FederatedPruning} framework to incorporate the same unstructured pruning technique used in our FedMap methodology. 
This adaptation, which we designate as \textit{FederatedPruning-UnstructuredLAMP}, employs the LAMP method \cite{LAMP}. 
Our comparative analysis, depicted in Figure \ref{fig:federated_pruning_experiments}, yields several insights: The iterative update of pruning masks at each specified interval $\mathbf{s}$ induces a notable drawdown effect in the performance of the baseline FederatedPruning method, which is proposed utilising structured pruning. 
This phenomenon can be attributed to significant portions of the model being alternately activated or deactivated across subsequent FL training cycles, necessitating a period of adjustment for the global model to realign with the overarching training objectives.

This initial performance setback is significantly mitigated when FederatedPruning is paired with our unstructured LAMP modification, and the drawdown effect is predominantly observed mostly at higher pruning thresholds. 
In more aggressive pruning scenarios, \textit{FederatedPruning}, using the unstructured approach, demonstrably surpasses FedMap in performance.

Conversely, FedMap is largely immune to the performance fluctuations associated with rapid changes in pruning masks. 
This stability is due to its pruning strategy, wherein the indices of parameters retained after each pruning event are strictly subsets of those selected in the preceding interval. 
This ensures consistent performance, with notable drawbacks only manifesting at exceptionally high pruning ratios. 
For instance, ResNet56 trained on the CIFAR100 dataset illustrates that while performance may decline at extreme pruning levels, it stabilises without further deterioration, unlike the fluctuating recovery observed in \textit{FederatedPruning}.
Across all phases of the FL training process, FedMap consistently demonstrates superior efficiency in reducing communication overhead compared to \textit{FederatedPruning}.

In summary, our FedMap methodology presents a formidable alternative to conventional approaches like \textit{FederatedPruning} \cite{FederatedPruning}. 
It strikes a delicate balance between optimising communication efficiency and maximising the performance benefits derived from prolonged model training. 
This comparison underscores a fundamental trade-off between minimising communication demands and leveraging extended training periods to enhance model performance without incurring adverse effects.

\begin{itemize}
    \item In centralised, i.i.d settings, Iterative Magnitude-based pruning achieves very high sparsity ratios while still retaining good performance. FedMap achieved convergence on par with FedAvg, but the sparsity ratios observed with non-federated (centralised) iterative pruning are not achievable, even in iid settings. 
    \item While FederatedPruning retains better performance than FedMap for very high sparsity, it never converges to the same performance as does FedMap.
    \item Another drawback of FederatedPruning is the high variance in model performance. While in FedMap, the client-models achieve high performance even at higher pruning ratios, the client-models in FederatedPruning wildly vary in performance, not guaranteeing stable performance.
    \item One additional benefit of FedMap is the relatively smaller transmission overhead, compared to FederatedPruning. While with FedMap, no masking information is ever transmitted, FederatedPruning transmits masking data every FL round during the pruning- and fine-tuning phase of its lifecycle.
\end{itemize}

\section{Conclusions, Limitations, and Future Work}
\label{sec:concl_lim_fut}
In this paper we introduced FedMap, a novel communication-efficient pruning technique tailored for federated learning deployments. Federated learning, while enabling privacy-preserving distributed training on decentralized data, often faces challenges due to the resource constraints of client devices. 
FedMap addresses this issue by collaboratively learning an increasingly sparse global model through iterative, unstructured pruning.
A key strength of FedMap is its ability to train a global model from scratch, making it well-suited for privacy-critical domains where suitable pretraining data are limited. 
By adapting iterative magnitude-based pruning to the federated setting, FedMap ensures that all clients prune and refine the same subset of the global model parameters, gradually reducing the model size and communication overhead.
The iterative nature of FedMap, where subsequent models are formed as subsets of predecessors, avoids parameter reactivation issues encountered in prior work, resulting in stable performance. 
Through extensive evaluation across diverse settings, datasets, model architectures, and hyperparameters, assessing performance in both IID and non-IID environments, we demonstrated the effectiveness of FedMap.
Comparative analysis against baseline approaches highlighted FedMap's ability to achieve more stable client model performance and superior results. 

During our evaluation, some limitations were identified, which, if addressed, could further enhance the capabilities and applicability of FedMap. 
One limitation lies in the current approach where all clients follow the same pruning schedule, using the same parameters $s$ and $p_G$. 
While this uniform approach simplifies the implementation, it may not be optimal for scenarios with heterogeneous client devices and varying resource constraints.
To address this limitation, future work could explore adaptive pruning schedules tailored to individual clients. 
By allowing clients to prune to varying degrees based on their specific bandwidth limitations or hardware capabilities, FedMap could better accommodate diverse client environments. 
Clients with limited bandwidth could adopt more aggressive pruning schedules to reduce communication overhead, while clients with more powerful hardware could maintain higher model complexities to maximize performance.

While methods like \textit{FederatedPruning} promise hardware-acceleration and thus increasing training speeds while reducing inference-latency, we focus on communication-reduction in this work and show that not only does iterative pruning lead to more stable model behaviour but also allows to retain higher model-performance at higher pruning rates.
Moreover, although the present study concentrates on the federated learning scenario, investigating the potential application of FedMap in other distributed learning paradigms, such as split learning or collaborative learning, could further broaden its impact and utility.
Overall, FedMap presents a promising solution to alleviate communication bottlenecks in federated learning systems while retaining model accuracy, offering a valuable contribution to the field of privacy-preserving distributed machine learning.

\bibliographystyle{IEEEtran}
\bibliography{bib.bib}

\newpage
\onecolumn

\appendices
\section{Additional Experimental Results}
\label{Appendix}
\subsection{FedMap Experimental Results}
\label{FedMapExpsAppendix}

\begin{figure*}[h!]
    \centering
    \begin{minipage}{.49\textwidth}
        \centering
        \textbf{Resnet56}
    \end{minipage}
    \begin{minipage}{.49\textwidth}
        \centering
        \textbf{MobilenetV2}
    \end{minipage}

    \vspace{0.5em}
    
    \begin{subfigure}[b]{\textwidth}
        \centering
      \includegraphics[width=.495\textwidth]{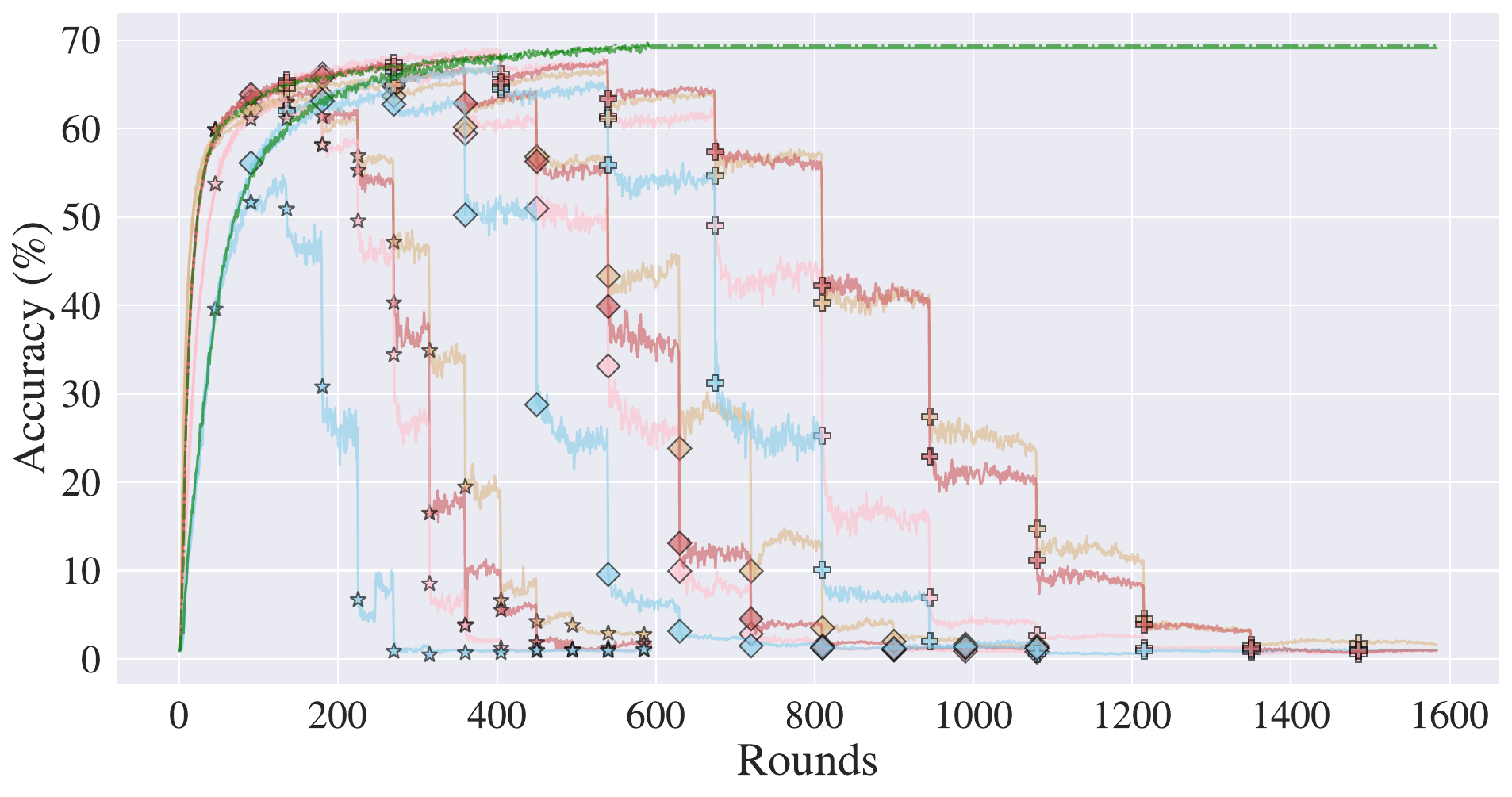}
      \includegraphics[width=.495\textwidth]{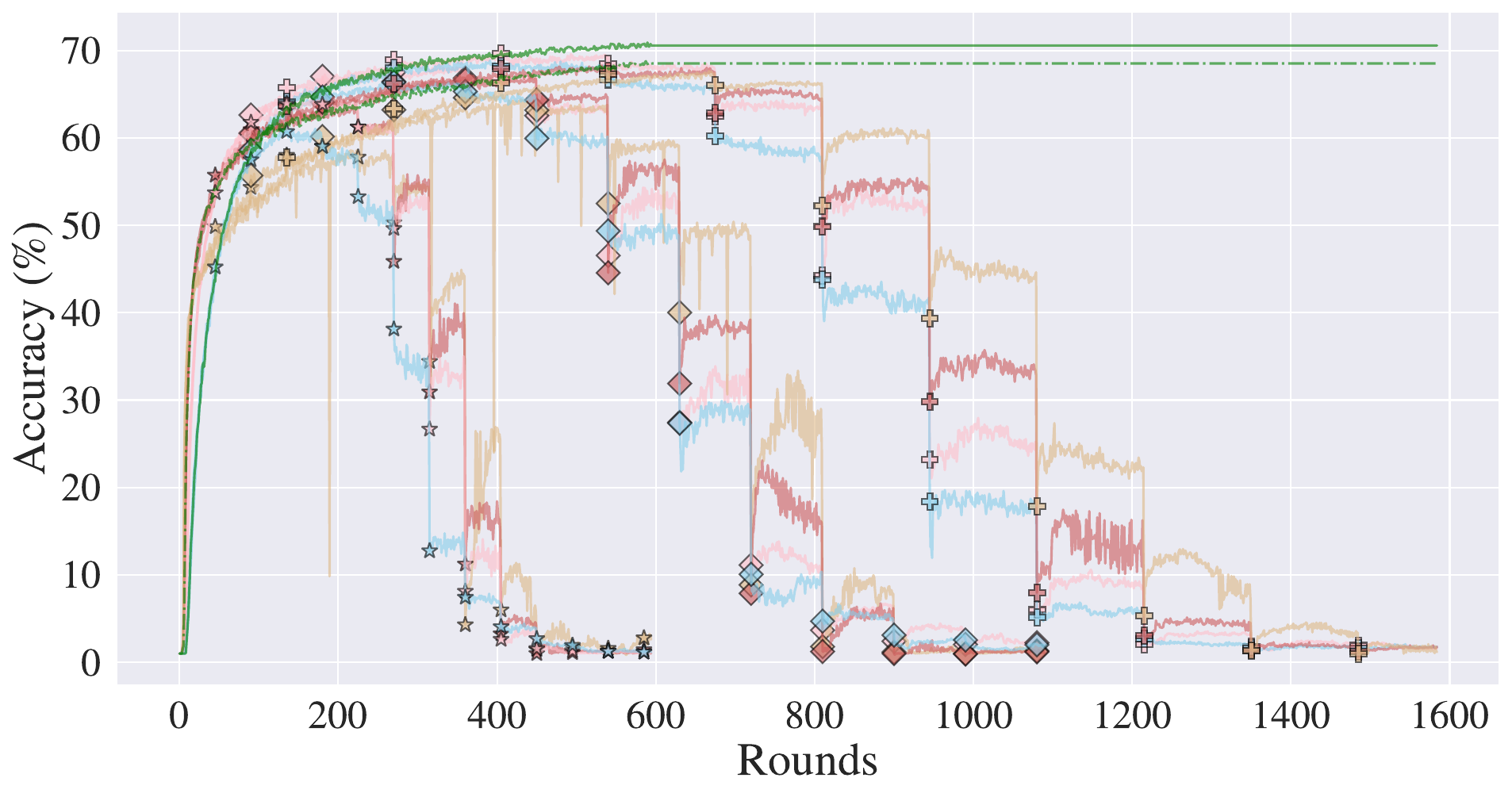}
      \vspace{-0.5em}
      \caption{CIFAR100}
      \label{fig:fedmap_cifar100_acc_rnds}
    \end{subfigure}%

    \vspace{1em}
    
    \begin{subfigure}[b]{\textwidth}
        \centering
      \includegraphics[width=.495\textwidth]{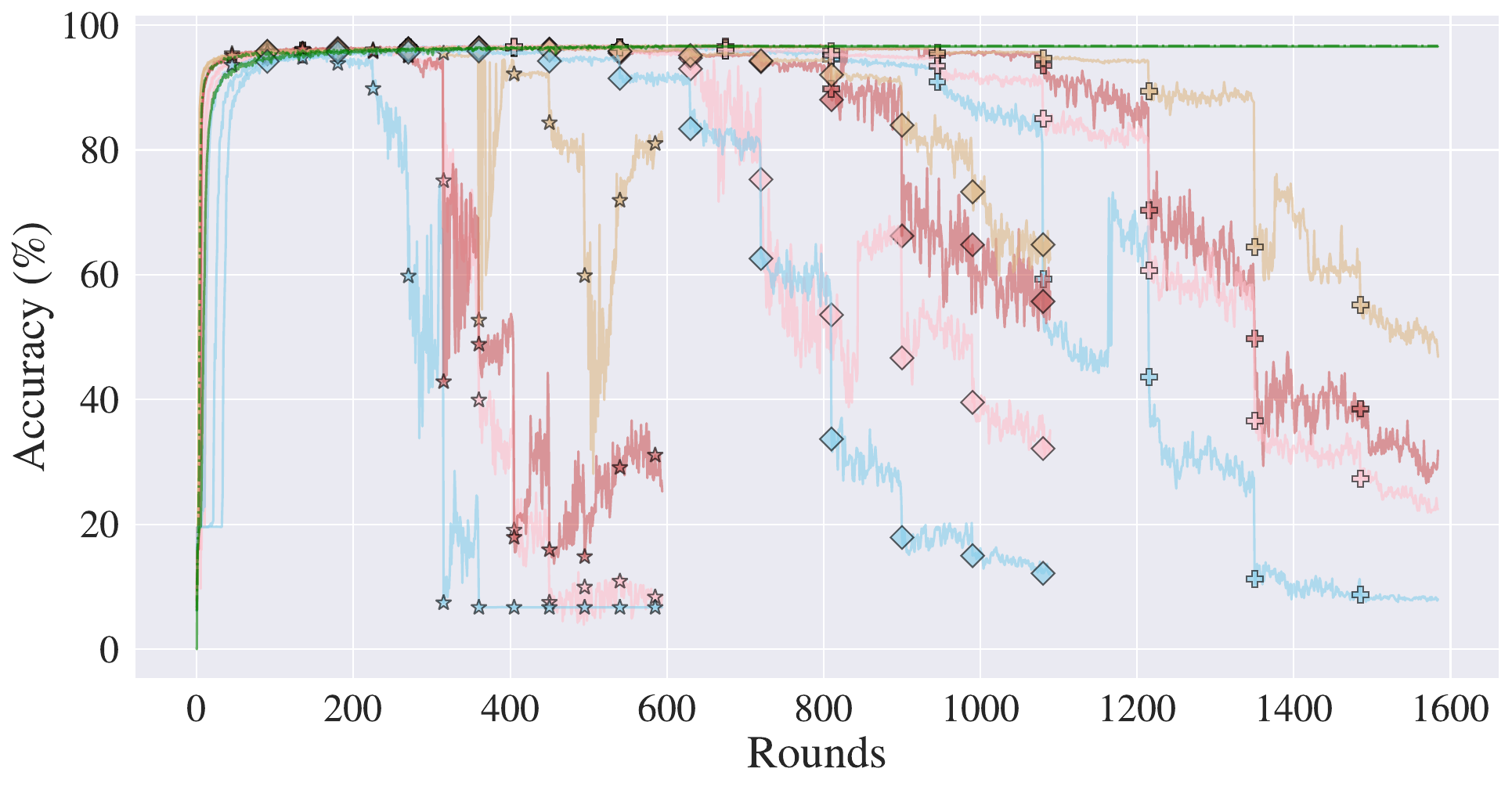}
      \includegraphics[width=.495\textwidth]{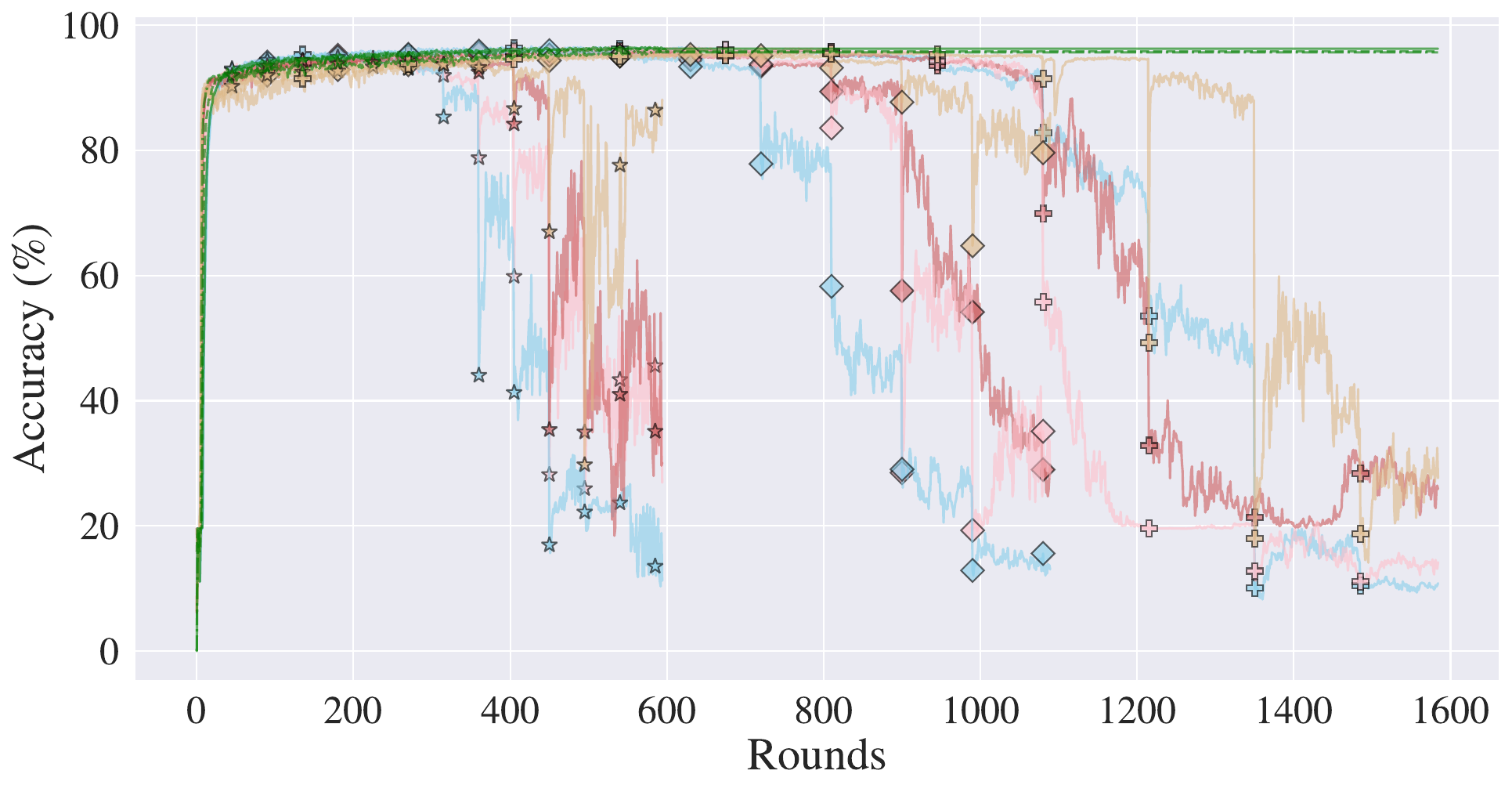}
      \vspace{-0.5em}
      \caption{SVHN}
      \label{fig:fedmap_svhn_acc_rnds}
    \end{subfigure}%

    \includegraphics[width=\textwidth, trim={0 7cm 0 7cm},clip]{figs/experiment_2/legend.pdf}

  \caption{Accuracy vs. rounds across variable number of local epochs ($L$) and step-widths ($s$). Experiments were performed with the following sets of hyperparameters: $L\in \{2,4,8\}, s\in \{45,90,135\}$}
  \label{fig:fedmap_main_results_appendix}
\end{figure*}

\begin{figure*}[ht]
    \includegraphics[width=1\textwidth]
    {figs/experiment_2/bars_legend.pdf}
    

    
    \begin{subfigure}[b]{\textwidth}
        \centering
      \includegraphics[width=\textwidth, trim={0 3em 0 0},clip]{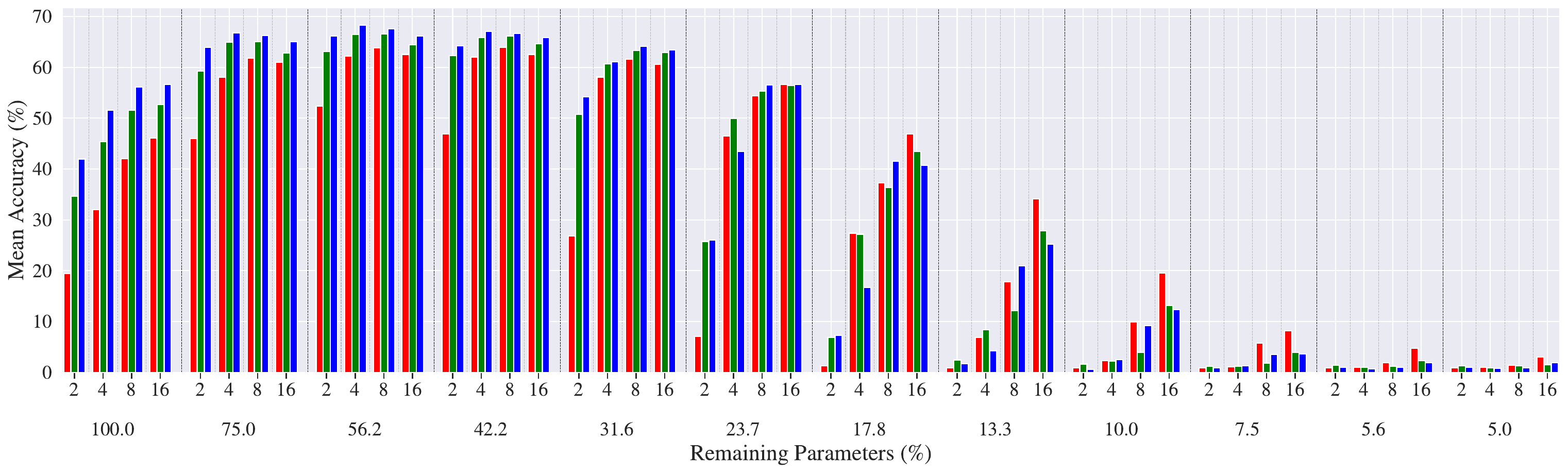}
      \vspace*{-8.5ex}  
        \begin{flushleft}
        \tiny
        \hspace{1.5em}$\mathbf{L}$\\
         \vspace{-0.5em}
        \dotfill \\
        Remaining \\
        Parameters \\
        \hspace{1.5em}(\%)
        \end{flushleft}
      \caption{CIFAR100}
      \label{fig:cifar100_resnet_bars}
    \end{subfigure}

    \begin{subfigure}[b]{\textwidth}
        \centering
        \includegraphics[width=\textwidth, trim={0 3em 0 0},clip]{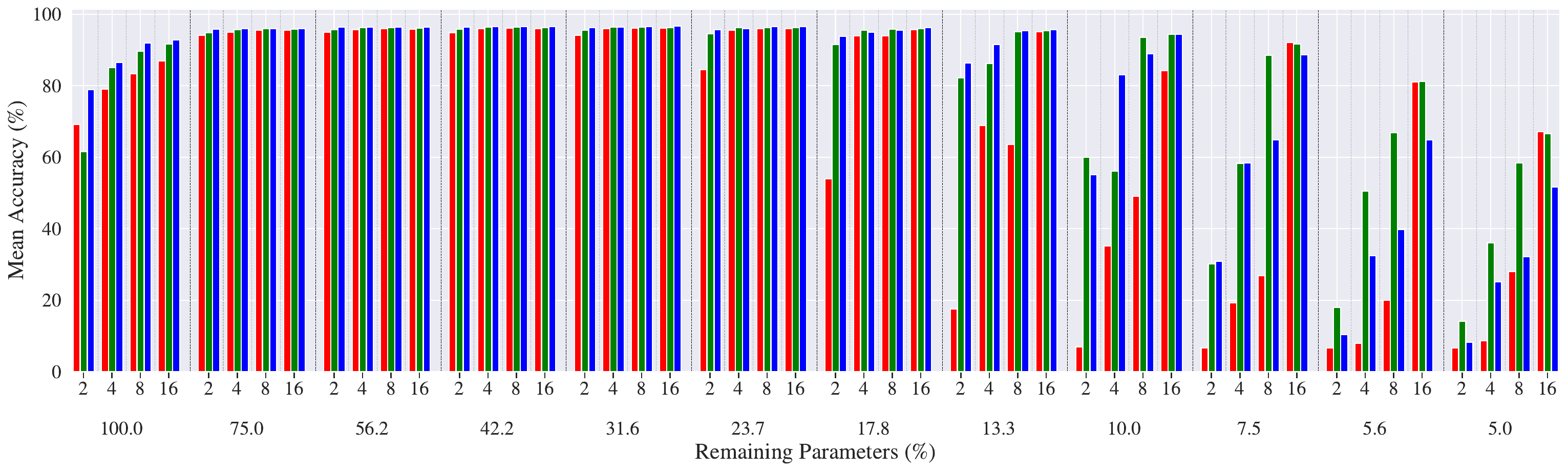}
        \vspace*{-8.5ex}  
        \begin{flushleft}
        \tiny
        \hspace{1.5em}$\mathbf{L}$\\
         \vspace{-0.5em}
        \dotfill \\
        Remaining \\
        Parameters \\
        \hspace{1.5em}(\%)
        \end{flushleft}
        \caption{SVHN}
         \label{fig:svhn_resnet_bars}
    \end{subfigure}
    \caption{Average performance per pruning step ($s$) across variable different values of local epochs ($L$) for Resnet56.}
    \label{fig:svhn_cifar100_resnet_bars}
\end{figure*}

\begin{figure*}[ht]
    \includegraphics[width=1\textwidth]
    {figs/experiment_2/bars_legend.pdf}



    
    \begin{subfigure}[b]{\textwidth}
        \centering
      \includegraphics[width=\textwidth, trim={0 3em 0 0},clip]{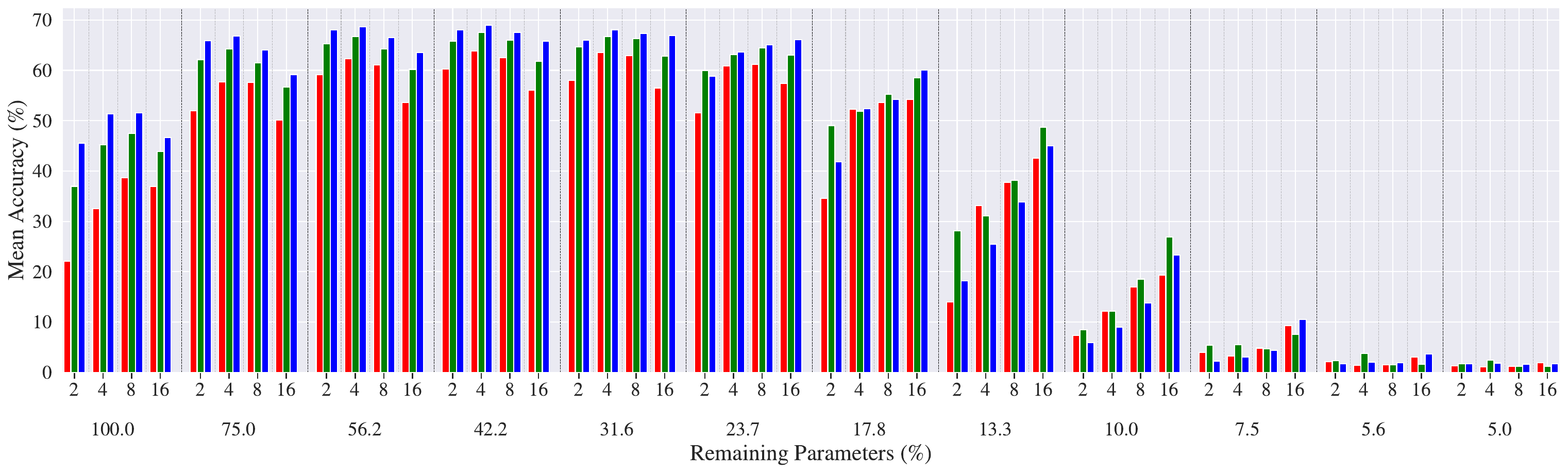}
      \vspace*{-8.5ex}  
        \begin{flushleft}
        \tiny
        \hspace{1.5em}$\mathbf{L}$\\
         \vspace{-0.5em}
        \dotfill \\
        Remaining \\
        Parameters \\
        \hspace{1.5em}(\%)
        \end{flushleft}
      \caption{CIFAR100}
      \label{fig:cifar_100_mobilnet_bars}
    \end{subfigure}

    \vspace{1em}
    
    \begin{subfigure}[b]{\textwidth}
        \centering
        \includegraphics[width=\textwidth, trim={0 3em 0 0},clip]{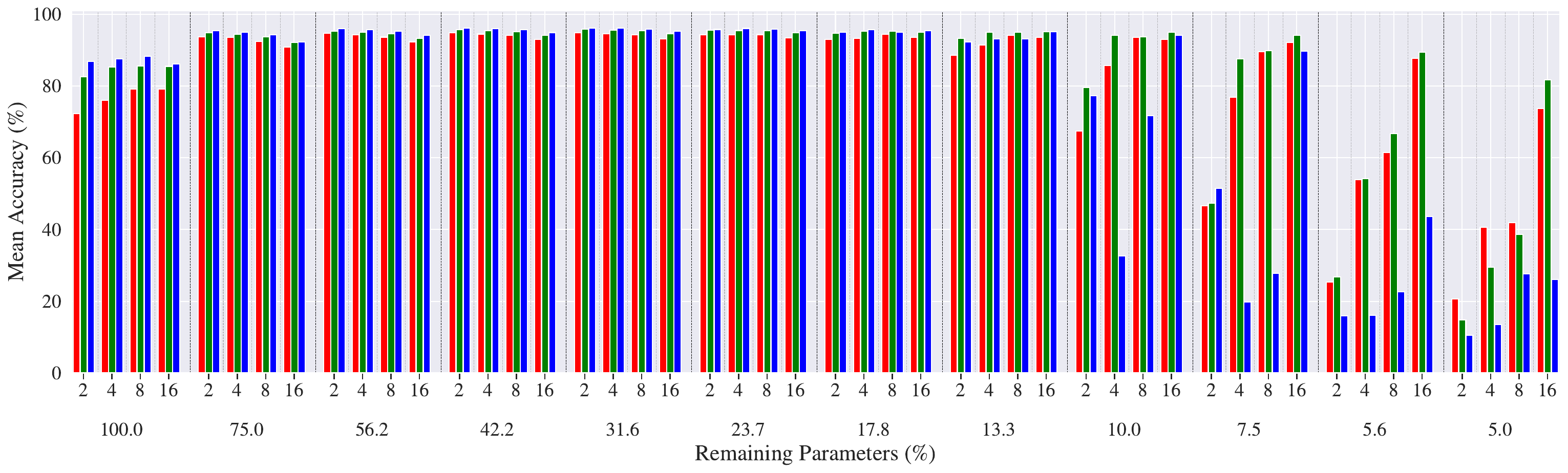}
        \vspace*{-8.5ex}  
        \begin{flushleft}
        \tiny
        \hspace{1.5em}$\mathbf{L}$\\
         \vspace{-0.5em}
        \dotfill \\
        Remaining \\
        Parameters \\
        \hspace{1.5em}(\%)
        \end{flushleft}
        \caption{SVHN}
        \label{fig:svhn_mobilenet_bars}
        \end{subfigure}

    \caption{Average performance per pruning step (s) across variable different values of local epochs (L) for MobileNetV2.}
    \label{fig:svhn_cifar100_mobilenet_bars}
\end{figure*}
\begin{figure*}[t]
        \begin{subfigure}[b]{.49\textwidth}
            \centering
          \includegraphics[width=\linewidth]{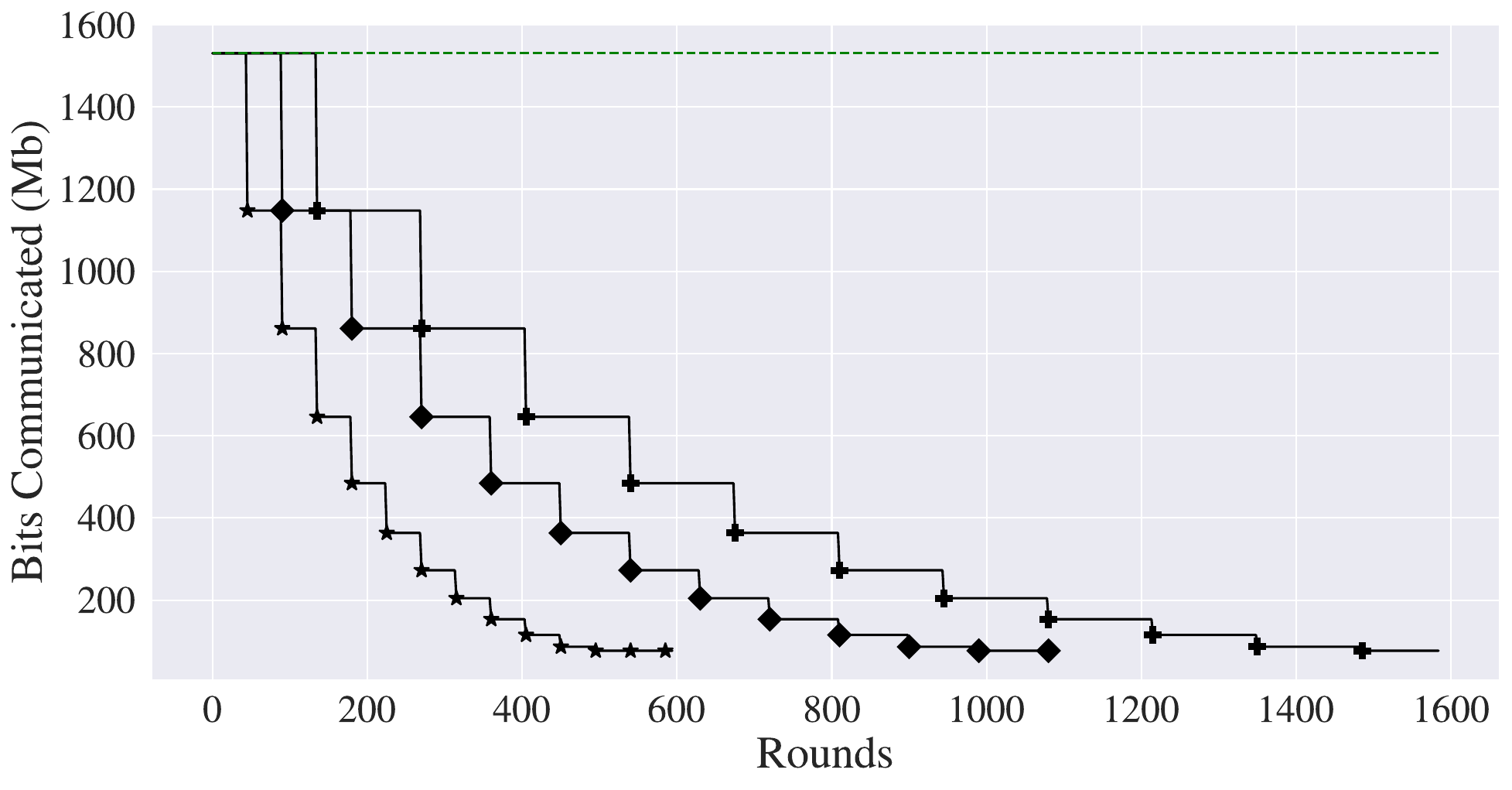}
          \caption{CIFAR100}
        \end{subfigure}
        \begin{subfigure}[b]{.49\textwidth}
            \centering
          \includegraphics[width=\linewidth]{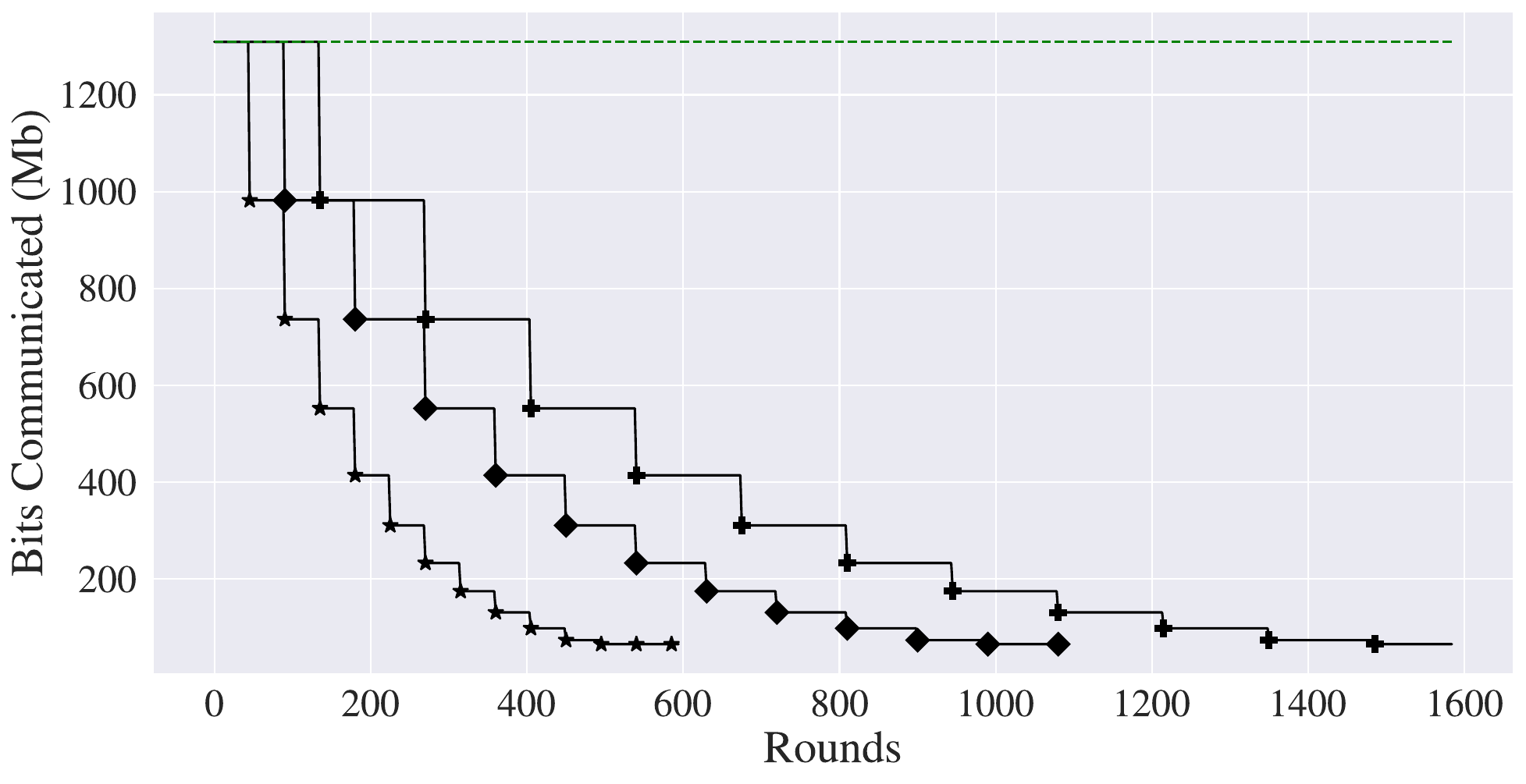}
          \caption{SVHN}
        \end{subfigure}    

     \includegraphics[width=1\textwidth]{figs/experiment_2/comms_legend.pdf}
     
    \caption{Communicated bits (Mb) per round of FL for exp. B. on MobileNetV2.}
    \label{fig:fedmap_main_comms_mobilenetv2}
\end{figure*}

\end{document}